\documentclass[twoside,11pt]{iopart}

\usepackage[utf8]{inputenc}
\usepackage{graphicx}
\usepackage{diagbox} 
\usepackage{hyperref}
\usepackage{iopams}
\usepackage{epstopdf}
\usepackage{txfonts}
\usepackage{url}

\usepackage{amsfonts}
\usepackage{amssymb} 
\usepackage[USenglish]{babel}
\usepackage{multirow}
\usepackage{paralist}
\usepackage{booktabs}
\usepackage{microtype}      
\usepackage{xcolor,colortbl}
\usepackage{hhline}
\usepackage{bbm}
\usepackage{tabularx}
\usepackage{lscape}
\usepackage{multirow,makecell}
\usepackage{color,soul}
\usepackage{algorithm}
\usepackage{algorithmicx}
\usepackage{algpseudocode}
\usepackage{xspace}
\usepackage{wrapfig}
\usepackage{harpoon}
\usepackage{mathrsfs}
\usepackage{bm}

\newtheorem{example}{Example}
\graphicspath{{figs/}}

\def\text{\textrm}
\def\dd{\mbox{d}}

\def\f{\frac}

\def\S{{\bf S}}

\def\P{{\bf P}}

\def\ie{\textit{i.e.}}
\def\eg{\textit{e.g.}}

\newcolumntype{R}[1]{>{\raggedleft\arraybackslash}p{#1}}

\definecolor{green}{rgb}{0.0, 0.6, 0.0}

\def\pro{\textcolor{green}{{\bf +}}}
\def\con{\textcolor{red}{{\bf --}}}

\begin{document}

\title[Spectrally Adapted Neural Networks]{Spectrally Adapted
  Physics-Informed Neural Networks for Solving Unbounded Domain
  Problems}
\author{Mingtao Xia$^1$, Lucas B\"{o}ttcher$^2$, Tom Chou$^1$}
\address{{\small $^1$ Dept.~of Mathematics, UCLA, Los Angeles, CA 90095-1555, USA}\\
{\small $^2$ Dept.~of Computational Science and Philosophy, Frankfurt School of Finance and Management, Frankfurt am Main, 60322, Germany}}
\ead{xiamingtao97@ucla.edu, l.boettcher@fs.de, tomchou@ucla.edu}

\date{\today}
\begin{abstract}
Solving analytically intractable partial differential equations (PDEs)
that involve at least one variable defined on an unbounded domain
arises in numerous physical applications.
%
%
%
Accurately solving unbounded domain PDEs requires efficient numerical
methods that can resolve the dependence of the PDE on the unbounded
variable over at least several orders of magnitude.
%
%
We propose a solution to such problems by combining two classes of
numerical methods: (i) adaptive spectral methods and (ii)
physics-informed neural networks (PINNs).  The numerical approach that
we develop takes advantage of the ability of physics-informed neural
networks to easily implement high-order numerical schemes to
efficiently solve PDEs and extrapolate numerical solutions at any
point in space and time.  We then show how recently introduced
adaptive techniques for spectral methods can be integrated into
PINN-based PDE solvers to obtain numerical solutions of unbounded
domain problems that cannot be efficiently approximated by standard
PINNs. Through a number of examples, we demonstrate the advantages of
the proposed spectrally adapted PINNs in solving PDEs and estimating
model parameters from noisy observations in unbounded domains.
\end{abstract}
%
%


\section{Introduction}
The use of neural networks as universal function
approximators~\cite{hornik1991approximation,park2020minimum} led to
various applications in
simulating~\cite{raissi2019physics,karniadakis2021physics} and
controlling~\cite{asikis2020neural,bottcher2021implicit,bottcher2022near,lewis2020neural}
physical, biological, and engineering systems.  Training neural
networks in function-approximation tasks is typically realized in two
steps. In the first step, an observable $u_{s}$ associated with each
distinct sample or measurement point $(x,t)_{s} \equiv (x_{s}, t_{s}),
\, s=1,2,\ldots, n$
%
%
is used to construct the corresponding loss function (\eg, the mean
squared loss) in order to find representations for the constraint $u_s
\equiv u(x_s, t_s)$ or infer the equations $u$ obeys.
%
%
In many physical settings, the variables $x$ and $t$ denote the space
and time variables, respectively. Thus, the data points $(x,t)_{s}$ in
many cases can be classified in two groups, $\{x_{s}\}$ and
$\{t_{s}\}$, and the information they contain may be manifested
differently in an optimization process. In the second step, the loss
function is minimized by backpropagating gradients to adjust neural
network parameters $\Theta$. If the number of observations $n$ is
limited, additional constraints may help to make the training process
more effective~\cite{DBLP:journals/corr/abs-1710-10686}.

To learn and represent the dynamics of physical systems, the
  constraints used in physics-informed neural networks
  (PINNs)~\cite{raissi2019physics,karniadakis2021physics} provide one
  possible option of an inductive bias in the training process. The
key idea underlying PINN-based training is that the constraints
imposed by the known equations of motion for some parts of the system
are embedded in the loss function. Terms in the loss function
associated with the differential equation can be evaluated using a
neural network, which could be trained via backpropagation and
automatic differentiation. In accordance with the distinction between
Lagrangian and Hamiltonian formulations of the equations of motion in
classical mechanics, physics-informed neural networks can be also
divided into these two
categories~\cite{lutter2019deep,roehrl2020modeling,zhong2019symplectic}.
Another formulation of PINNs uses variational
principles~\cite{kharazmi2019variational} in the loss function to
further constrain the types of functions used. Such variational PINNs
rely on finite element (FE) methods to discretize partial differential
equation (PDE)-type constraints.

Many other PINN-based numerical algorithms have been recently
proposed. A space-time domain decomposition PINN method was proposed
for solving nonlinear PDEs~\cite{jagtap2020extended}. In other
variants, physics-informed Fourier neural operators have also been
proposed to learn the underlying PDE models~\cite{li2021physics}. In
general, PINNs link modern neural network methods with traditional
complex physical models and allow algorithms to efficiently use
higher-order numerical schemes to (i) solve complex physical problems
with high accuracy, (ii) infer model parameters, and (iii) reconstruct
physical models in data-driven inverse
problems~\cite{raissi2019physics}. Therefore, PINNs have become
increasingly popular as they can avoid certain computational
difficulties encountered when using traditional FE/FD methods to find
solutions to physics models.
%
%

The broad utility of PINNs is reflected in their application to
aerodynamics~\cite{mao2020physics}, surface
physics~\cite{fang2019physics}, power
systems~\cite{misyris2020physics}, cardiology~\cite{sahli2020physics},
and soft biological tissues~\cite{liu2020generic}. When implementing
PINN algorithms to find functions in an unbounded system, the
unbounded variables cannot be simply normalized, precluding
reconstruction of solutions outside the range of data. Nonetheless,
many problems in nature are associated with long-ranged
potentials~\cite{bottcher2021computational,strub2019modeling} (\ie,
unbounded spatial domains) and processes that are subject to algebraic
damping~\cite{barre2011algebraic} (\ie, unbounded temporal domains),
and thus need to be solved in unbounded domains. For example, to
capture the oscillatory and decaying behavior at infinity of the
solution to Schr\"odinger's equation, efficient numerical methods are
required in the unbounded domain $\mathbb{R}$ \cite{li2018stability}.
As another example, in structured cellular proliferation models in
mathematical biology, efficient unbounded domain numerical methods are
required to detect and better resolve possible blow-up in mean cell
size ~\cite{xia2020pde,xia2021kinetic}. Finally, in solid-state
physics, long-range interactions
\cite{mengotti2011real,hugli2012artificial} require algorithms
tailored for unbounded domain problems to accurately simulate particle
interactions over long distances.

Solving unbounded domain problems is thus a key challenge in various
fields that cannot be addressed with standard PINN-based solvers. To
efficiently solve PDEs in unbounded domains, we will treat the
information carried by the $x_{s}$ data using spectral decompositions
of $u$ in the $x$ variable. Typically, a spatial initial condition of
the desired solution is given and some spatial regularity is assumed
from the underlying physical process. As a consequence, we suppose
that we can use a spectral expansion in $x$ to record spatial
information. On the other hand, a solution's behavior in time $t$ is
unknown and one still has to numerically step forward in time to
obtain the solution. Thus, we combine PINNs with spectral methods and
propose a spectrally adapted PINN (s-PINN) method that can also
utilize recently developed adaptive function expansions
techniques~\cite{xia2021efficient,xia2021frequency}.

In contrast to traditional numerical spectral schemes that can only
furnish solutions at discrete, predetermined timesteps, our approach
uses time $t$ as an input variable into the neural network combined
with the PINN method to define a loss function, which enables (i) easy
implementation of high-order Runge-Kutta schemes to relax the
constraint on timesteps and (ii) easy extrapolation of the numerical
solution at any time.  However, our approach is distinct from that
taken in standard PINN, variational-PINN, or physics-informed neural
operator approaches.  We do not input $x$ into the network or try to
learn $u(x)$ as a composition of, \eg, Fourier neural operators;
instead, we assume that the function can be approximated by a spectral
expansion in $x$ with appropriate basis functions. Rather than
learning the explicit spatial dependence directly, we train the neural
network to learn the time-dependent expansion coefficients. Our main
contributions include (i) integrating spectral methods into
multi-output neural networks to approximate the spectral expansions of
functions when partial information is available, (ii) incorporating
recently developed adaptive spectral methods in our s-PINNs, and (iii)
presenting explicit examples illustrating how s-PINNs can be used to
solve unbounded domain problems, recover spectral convergence, and
more easily solve inverse-type PDE inference problems.
%
%
We show how s-PINNs provide a unified, easy-to-implement method for
solving PDEs and performing parameter-inference given noisy
observation data and how complementary adaptive spectral techniques
can further improve efficiency, especially for solving problems in
unbounded domains.

In Sec.~\ref{spectral_network}, we show how neural networks can be
combined with modern adaptive spectral methods to outperform standard
neural networks in function approximation tasks. As a first
application, we show in Sec.~\ref{pde_solving} how efficient PDE
solvers can be derived from spectral PINN methods. In
Sec.~\ref{model_reconstruction}, we discuss another application that
focuses on reconstructing underlying physical models and inferring
model parameters given observational data. In Sec.~\ref{summary}, we
summarize our work and discuss possible directions for future
research. A summary of the main variables and parameters used in this
study is given in Table~\ref{tab:model_variables}. Our source codes
are publicly available at
\url{https://gitlab.com/ComputationalScience/spectrally-adapted-pinns}.

\begin{table}[htb]
  \caption{\textbf{Overview of variables.} Definitions of the main
    variables and parameters used in this paper.}
  \vspace{2mm}
\renewcommand*{\arraystretch}{1.0}
\begin{tabular}{| >{\centering\arraybackslash} m{4em}| 
>{\arraybackslash} m{30em}|}\hline
Symbol & Definition
\\[1pt] \hline\hline
   \,\,\, $n$\,\, & number of observations\\[2pt]  \hline
    \,\,\, $N$\,\, & spectral expansion order \\[2pt]  \hline
   \,\,\, $N_H$\,\, & number of intermediate layers in the neural network \\[2pt] \hline
     \,\,\, $H$\,\, & number of neurons per layer \\[2pt]  \hline
     \,\,\, $\eta$\,\, & learning rate of stochastic gradient descent \\[2pt]  \hline
  \,\,\, $\Theta$\,\, & neural network hyperparameters \\[2pt]  \hline
   \,\,\, $K$\,\, & order of the Runge--Kutta scheme \\[2pt]  \hline
   \,\,\, $\beta$\,\, & scaling factor of basis functions $\phi_{i, x_L}^{\beta}(x)\coloneqq\phi_i(\beta(x-x_L))$\\[2pt]  \hline
      \,\,\, $x_L$\,\, & translation of basis functions 
$\phi_{i, x_L}^{\beta}\coloneqq\phi_i(\beta(x-x_L))$ \\[2pt]  \hline
        \,\,\, $u_{N, x_L}^{\beta}$\,\, & spectral expansion of order 
$N$ generated by the neural network: $u_{N, x_L}^{\beta}=\sum_{i=0}^N w_{i, x_L}^{\beta}\phi_i(\beta(x-x_L))$ \\[2pt]  \hline
  \,\,\, $\mathcal{F}(u_{N, x_L}^{\beta})$\,\, & frequency indicator for the spectral expansion $u_{N, x_L}^{\beta}$ \\[2pt]  \hline
 \,\,\, $\hat{\mathcal{H}}_{i, x_L}^{\beta}$\,\, & generalized Hermite function of order $i$, 
scaling factor $\beta$, and translation $x_L$ \\[2pt]  \hline
    \,\,\, $P_{N, x_L}^{\beta}$\,\, & function space defined by 
the first $N+1$ generalized Hermite functions 
$P_{N, x_L}^{\beta}\coloneqq\{\hat{\mathcal{H}}_{i, x_L}^{\beta}\}_{i=0}^N$ \\[2pt]  \hline
 \,\,\, $q$\,\, & scaling factor ($\beta$) adjustment ratio \\[2pt]  \hline
 \,\,\, $\nu$\,\, & threshold for adjusting the scaling factor $\beta$ \\[2pt]  \hline
 \,\,\, $\rho, \rho_0$\,\, & threshold for increasing, decreasing $N$ \\[2pt]  \hline
  \,\,\, $\gamma$\,\, & ratio for adjusting $\rho$ \\[2pt]  \hline
\end{tabular}
\label{tab:model_variables}
\end{table}
\section{Combining Spectral Methods with Neural Networks}
\label{spectral_network}

\begin{figure}[h!]
    \begin{center}
\includegraphics[width=4.5in]{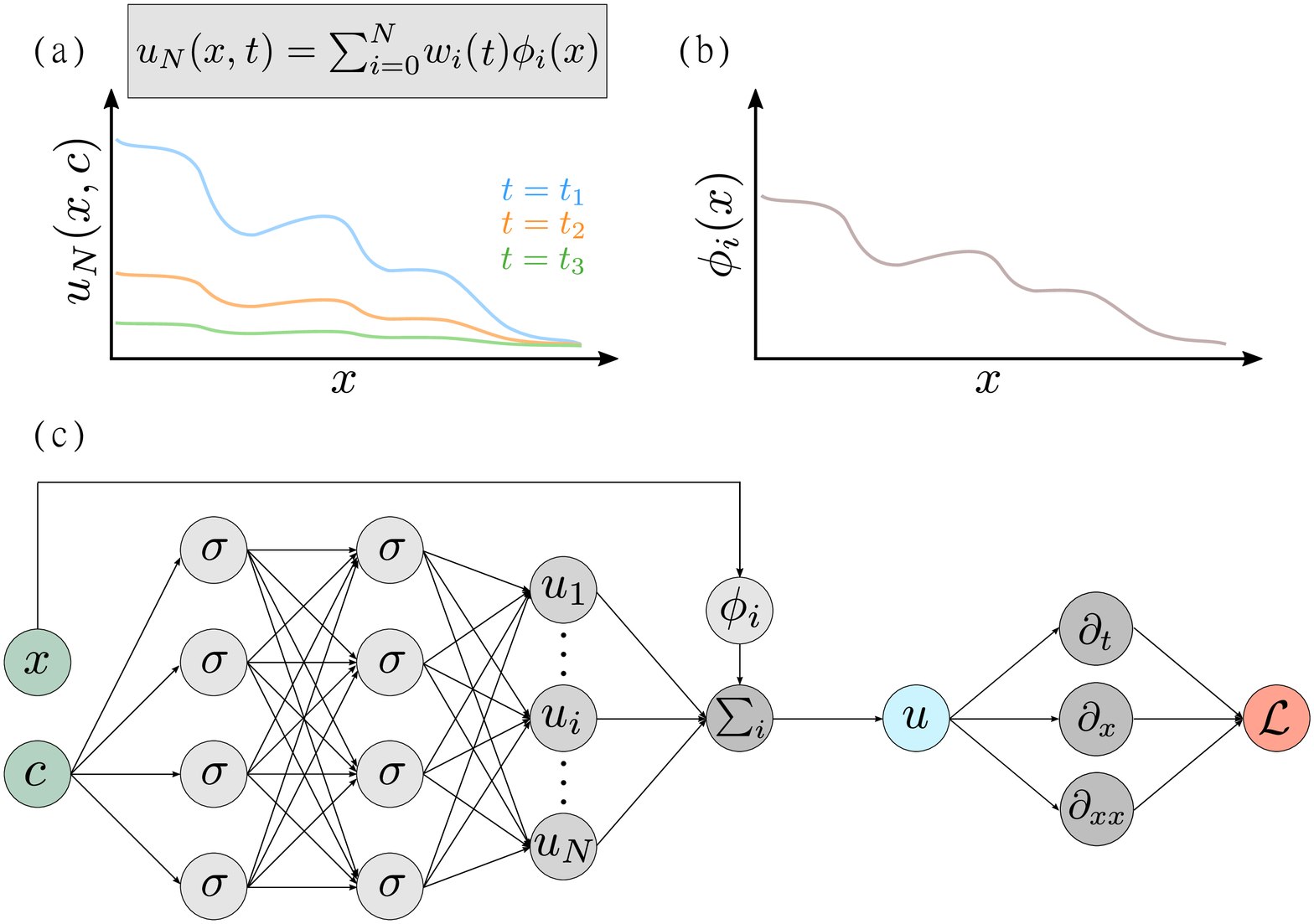}
    \caption{\small Solving unbounded domain problems with spectrally
      adapted physics-informed neural networks for functions $u_N(x,t)$
      that can be expressed as a spectral expansion
      $u_N(x,t)=\sum_{i=0}^N w_i(t)\phi_{i}(x)$.  (a) An example of a
      function $u_N(x,t)$ plotted at three different time points. (b)
      Decaying behavior of a corresponding basis function element
      $\phi_i(x)$. (c) PDEs in unbounded domains can be solved by
      combining spectral decomposition with the PINNs and minimizing
      the loss function $\mathcal{L}$. Spatial derivatives of
      basis functions are explicitly defined and easily obtained.}
    \label{fig:spectral_pinn}
\end{center}
\end{figure}

In this section, we first introduce the basic features of function
approximators that rely on neural networks and spectral methods
designed to handle variables that are defined in unbounded domains.
In a dataset $(x_s, t_s, u_{s})$, $s\in\{1,\dots,n\}$, $x_{s}$ are
values of the sampled ``spatial'' variable $x$ which can be defined
in an unbounded domain. We will also assume that our problem is
defined within a finite time horizon so that $t_{s}$ are time points
restricted to a bounded domain, and are thus normalizable. One
central goal is to approximate the constraint $u_s \coloneqq u(x_s,
t_s)$ by computing the function $u(x, t)$ and the equation it
obeys. Our key assumption is that the solution's behavior in $x$ can
be represented by a spectral decomposition, while $u$'s behavior in
$t$ remains unknown and is to be learned from the neural network. This
is achieved by isolating the possibly unbounded spatial variables
$x$ from the bounded variables $t$ by expressing $u$ in terms of
suitable basis functions in $x$ with time-dependent weights. As
indicated in Fig.~\ref{fig:spectral_pinn}(a), we approximate $u_s$
using
\begin{equation}
u_{s} \coloneqq u(x_s, t_s)\approx u_N(x_{s}, t_{s})\coloneqq  \sum_{i=0}^N 
w_i(t_s)\phi_i(x_s),
\label{spectral_approx}
\end{equation}
where $\{\phi_i\}_{i=0}^N$ are suitable basis functions that can be
used to approximate $u$ in an unbounded domain (see
Fig.~\ref{fig:spectral_pinn}(b) for a schematic of a basis function
$\phi_i(x)$ that decays with $x$).  Examples of such basis functions
include, for example, the generalized Laguerre functions in
$\mathbb{R}^+$ and the generalized Hermite functions in
$\mathbb{R}$~\cite{shen2011spectral}. In addition to being defined on
an unbounded domain, spectral expansions allow high
accuracy~\cite{trefethen2000spectral} calculations with errors that
decay exponentially (spectral convergence) in space if the target
function $u(x,t)$ is smooth.

Figure~\ref{fig:spectral_pinn}(c) shows a schematic of our proposed
spectrally adapted PINN algorithm.  The variable $x$ is directly fed
into the basis functions $\phi_i$ instead of being used as an input in
the neural network. If one wishes to connect the output
$u_N(x,t; \Theta)$ of the neural network to the solution of a
PDE, one has to include derivatives of $u$ with respect to $x$ and $t$
in the loss function $\mathcal{L}$.  Derivatives that involve the
variable $x$ can be easily and explicitly calculated by taking
derivatives of the basis functions with high accuracy while
derivatives with respect to $t$ can be obtained via automatic
differentiation~\cite{linnainmaa1976taylor,paszke2017automatic}.

If a function $u$ can be written in terms of a spectral expansion in
some dimensions (\textit{e.g.}, $x$ in Eq.~\ref{spectral_approx}) with
appropriate spectral basis functions, we can approximate $u$ using a
multi-output neural network by solving the corresponding least squares
optimization problem
\begin{eqnarray}
 \min_{\Theta}\left\{\sum_s \big\vert 
u_N(x_s, t_s;\Theta) - u_{s} \big\vert^2\right\},\,\,\,
u_N(x, t; \Theta) = \sum_{i=0}^N
    w_i(t;\Theta)\phi_i(x),
\label{uhat}
\end{eqnarray}
where $\Theta$ is the hyperparameter set of a neural network that
outputs the $t$-dependent vector of weights $w_i(t;\Theta)$.  This
representation will be used in the appropriate loss function depending
on the application. The neural network can achieve arbitrarily high
accuracy in the minimization of the loss function if it is deep enough
and contains sufficiently many neurons in each layer
\cite{hornik1989multilayer}. Since the solution's spatial behavior has
been approximated by the spectral expansion which could achieve high
accuracy with proper $\phi_i$, we shall show that solving
Eq.~\ref{uhat} can be more accurate and efficient than directly
fitting to $u_{s}$ by a neural network without using a spectral
expansion.

As a motivating example, we compare the approximation error of a
neural network which is fed both $x_{s}$ and $t_{s}$ with that of the
s-PINN method in which only $t_{s}$ are inputted, but with the
information contained in $x_{s}$ imposed on the solution via basis
functions of $x$. We show that taking advantage of the prior knowledge
on the $x$-data greatly improves training efficiency and accuracy. All
neural networks that we use in our examples are based on fully
connected linear layers with ReLU activation functions. Weights in
each layer are initially distributed according to a uniform
distribution $\mathcal{U}(-\sqrt{a},\sqrt{a})$, where $a$ is the
inverse of the number of input features. To normalize hidden-layer
outputs, we apply the batch normalization
technique~\cite{ioffe2015batch}. Neural-network parameters are
optimized using stochastic gradient descent.

\begin{example}\hspace{-5pt}{\bf: Function approximation}\\
\label{example_simulate}
\rm Consider approximating the function
\begin{equation}
u(x, t) = {8x \sin 3x \over 
\left(x^{2} + 4\right)^{2}}\, t,
\label{sim_target}
\end{equation}
which decays algebraically as $u(x\to \infty, t) \sim t/|x|^3$ when
$|x|\rightarrow\infty$. To numerically approximate
Eq.~\ref{sim_target}, we choose the loss function to be the
mean-squared error
\begin{equation}
{\rm MSE} = {1 \over n} \sum_{s=1}^{n} \big\vert u_{N}(x_s, t_s) - u_{s}\big\vert^{2}.
   \label{MSEerror}
\end{equation}

A standard neural network approach is applied by inputting \textit{both}
$x_{s}$ and $t_{s}$ into a 5-layer, 10 neuron-per-layer network
defined by hyperparameters $\tilde{\Theta}$ to find a numerical
approximation to $u_{N}(x_s,t_s)\coloneqq \tilde{u}(x_s,
t_s;\tilde{\Theta})$ by minimizing Eq.~\ref{MSEerror} with respect
to $\tilde{\Theta}$ (the $\tilde{u}, \tilde{\Theta}$ notation refers to
hyperparameters in the non-spectral neural network).

To apply a spectral multi-output neural network to this problem, we
need to choose an appropriate spectral representation of the spatial
dependence of Eq.~\ref{sim_target}, in the form of Eq.~\ref{uhat}.
In order to capture an algebraic decay at infinity as well as the
oscillatory behavior resulting from the $\sin(3x)$ term, we start from
the modified mapped Gegenbauer functions
(MMGFs)~\cite{tang2020rational}
\begin{equation}
R_i^{\lambda, \beta}(x)=
(1+(\beta{x})^2)^{-(\lambda+1)/2}C_i^{\lambda}\!
\left(\beta x/\sqrt{1+(\beta x)^{2}}\right),\,\,\, x\in\mathbb{R},
\end{equation}
where $C_i^{\lambda}(\cdot)$ is the Gegenbauer polynomial of order
$i$. At infinity, the MMGFs decay as $R_i^{\lambda, \beta}(x)\sim
\textrm{sign}(x)^i\frac{(2\lambda)^{(i)}}{i!}(1+(\beta{x})^2)^{-(\lambda+1)/2}$,
where $(2\lambda)^{(i)}$ is the $i^{\rm th}$ rising factorial of
$2\lambda$.  A suitable basis $\phi_{i}$ needs to include functions
that decay more slowly than $x^{-3}$. If we choose $\beta = 1/2$ and
the special case $\lambda=0$, the basis function is defined as
$\phi_{i}(x) = R_i^{0, \beta}(x)\equiv (1+(\beta
x)^{2})^{-1/2}T_{i}(\beta x/\sqrt{1+(\beta x)^{2}})$, where $T_{i}$
are the Chebyshev polynomials. We thus use
\begin{equation}
    u_{N}(x_{s}, t_{s}; \Theta) = \sum_{i=0}^{N=9}
    w_{i}(t_{s};\Theta)R_i^{0, \beta}(x_{s})
\end{equation}
in Eq.~\ref{MSEerror} and use a 4-layer neural network with 
10 neurons per layer to learn the coefficients
$\{w_i(t;\Theta)\}_{i=0}^9$ by minimizing the MSE (Eq.~\ref{MSEerror}) 
with respect to $\Theta$.
 \begin{figure}[t]
      \begin{center}
  \includegraphics[width=4.4in]{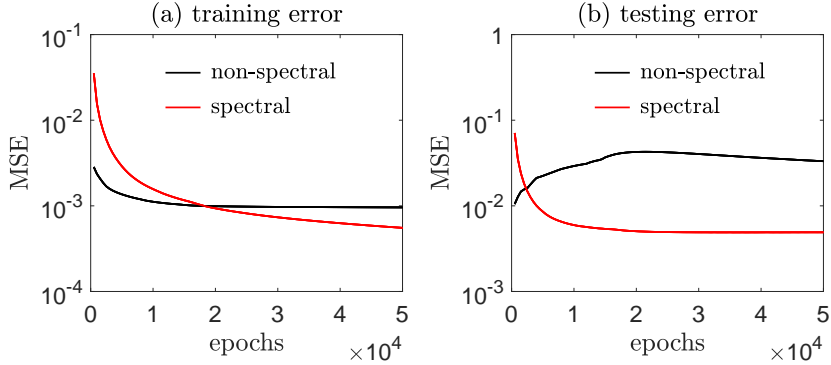}
	\end{center}
      \vspace{-4mm}
       \caption{\small Example~\ref{example_simulate}: Function
           approximation. Approximation of the target function
         Eq.~\ref{sim_target} using both standard neural networks
         and a spectral multi-output neural network that learns the
         coefficients $w_i(t;\Theta)$ in the spectral expansion
         Eq.~\ref{spectral_approx}.  Comparison of the approximation
         error using a spectral multi-output neural network (red) with
         the error incurred when using a standard neural-network
         function approximator (black).  Here, both the spectral and
         non-spectral function approximators use the same number of
         parameters, but the spectral multi-output neural network
         converges much faster on the training set and has a smaller
         testing error than the standard neural network. (a) The
         training curve of the spectral multi-output neural network
         decreases much faster than that of the standard neural
         network. (b) Since the spectral multi-output neural network
         is better at fitting the data by taking advantage of the
         spectral expansion in $x$, its testing error is also much
         smaller and decreases faster.}
     \label{fig_ex1}
\end{figure}
The total numbers of parameters for both the 4-layer spectral
multi-output neural network and the normal 5-layer neural network are
the same. The training set and the testing set each contain $n=200$
pairs of values $(x, t)_{s}=(x_{s}, t_{s})$ where $x_s$ are sampled
from the Cauchy distribution, $x_s \sim \mathcal{C}(12, 0)$, and $t_s
\sim \mathcal{U}(0, 1)$. For each pair $(x_{s}, t_{s})$, we find
$u_{s} = u(x_{s}, t_{s})$ using Eq.~\ref{sim_target}.  Clearly,
$x_s$ is sampled from the unbounded domain $\mathbb{R}$ and cannot be
normalized (the expectation and variance of the Cauchy distribution do
not exist).

We set the learning rate $\eta=5\times10^{-4}$ and plot the training
and testing MSEs (Eq.~\ref{MSEerror}) as a function of the number of
training epochs in Fig.~\ref{fig_ex1}.  Figures~\ref{fig_ex1}(a) and
(b) show that the spectral multi-output neural network yields smaller
errors since it naturally and efficiently captures the oscillatory and
decaying feature of the underlying function $u$ from
Eq.~\ref{sim_target}.  Directly fitting $u\approx\tilde{u}$ leads to
over-fitting on the training set which does nothing to reduce the
testing error. Therefore, it is important to take advantage of the
data structure, in this case, using the spectral expansion to
represent the function's known oscillations and decay as $x\to
\infty$. In this and subsequent examples, all computations are
performed using Python 3.8.10 on a laptop with a 4-core
Intel\textsuperscript{\textregistered} i7-8550U CPU @ 1.80 GHz.

\end{example}

%
\section{Application to Solving PDEs}
\label{pde_solving}
In this section, we show that spectrally adapted neural networks can
be combined with physics-informed neural networks (PINNs) which we
shall call spectrally adapted PINNs (s-PINNs). We apply s-PINNs to
numerically solve PDEs, and in particular, spatiotemporal PDEs in
unbounded domains for which standard PINN approaches cannot be
directly applied. Although we mainly focus on solving spatiotemporal
problems, s-PINNs are also applicable to other types of PDEs.

Again, we assume that the problem is defined over a finite
time horizon $t$ while the spatial variable $x$ may be defined in an
unbounded domain.  Assuming the solution's asymptotic behavior in $x$
is known, we approximate it by a spectral expansion in $x$ with
suitable basis functions (\textit{e.g.}, MMGFs in
Example~\ref{example_simulate} for describing algebraic decay at
infinity).  Assuming $\mathcal{M}$ is an operator that only involves
the spatial variable $x$ (\textit{e.g.}, $\partial_x,
\partial_{x}^{2}$, etc.), we can represent the solution to the
spatiotemporal PDE $\partial_{t}u = \mathcal{M}[u](x, t)$ by the
spectral expansion in Eq.~\ref{uhat} with expansion coefficients
$\{w_i(t;\Theta)\}$ to be learned by a neural network with
hyperparameters $\Theta$.
%
%
If the solution's behavior in both $x$ and $t$ are known and one can
find proper basis functions in both the $x$ and $t$ directions, then
one could use a spectral expansion in both $x$ and $t$ to solve the
PDE directly without time-stepping. However, it is often the case that
the time dependence is unknown and $u(x, t)$ needs to be solved
step-by-step in time.

As in standard PINNs, we use a high-order Runge--Kutta scheme to
advance time by uniform timesteps $\Delta t$. What distinguishes our
s-PINNs from standard PINNs is that only the intermediate times
$t_{s}$ between timesteps are defined as inputs to the neural network,
while the outputs contain global spatial information (the spectral
expansion coefficients), as shown in Fig.~\ref{fig:spectral_pinn}(c).
Over a longer time scale, the optimal basis functions in the spectral
expansion Eq.~\ref{uhat} may change. Therefore, one can use new
adaptive spectral methods proposed in
\cite{xia2021efficient,xia2021frequency}. Using s-PINNs to solve PDEs
has the advantages that they can (i) accurately represent spatial
information via spectral decomposition, (ii) convert solving a PDE
into an optimization and data fitting problem, (iii) easily implement
high-order, implicit schemes to advance time with high accuracy, and
(iv) allow the use of recently developed spectral-adaptive techniques
that dynamically find the most suitable basis functions.

The approximated solution to the PDE $\partial_{t}u =
\mathcal{M}[u](x, t)$ can be written at discrete timesteps
$t_{j+1}-t_{j} = \Delta t$ as
\begin{equation}
    u_N(x, t_{j+1}; \Theta_{j+1}) = \sum_{i=0}^N w_i(t_{j+1}; \Theta_{j+1})\phi_i(x),
    \label{numti}
\end{equation}
where $\Theta_{j+1}, j\geq 1$ is the hyperparameter set of the neural
network used in the time interval $(j\Delta t, (j+1)\Delta t)$.  In
order to forward time from $t_{j}=j\Delta t$ to $t_{j+1} = (j+1)\Delta
t$, we can use, \eg, a $K^{\text{th}}$-order implicit Runge--Kutta
scheme, with $0<c_s<1$ ($s=1,\ldots,K$) as parameters describing
different collocation points in time and $a_{rs}, b_r\,
(r=1,\ldots,K)$ the associated coefficients.

Given $u(x, t_j)$, the $K^{\text{th}}$-order implicit Runge--Kutta
scheme aims to approximate $u(x, t_{j}+c_s\Delta{t})$ and $u(x,
t_j + \Delta{t})$ through

\begin{eqnarray}
 u_N(x, t_j+c_s \Delta{t}) & = &
   u(x, t_j) + \sum_{r=1}^K a_{rs}
  \mathcal{M}\big[u_N(x, t_j + c_r\Delta{t})\big],  \nonumber \\
\quad u_N(x, t_j+\Delta{t}) &=& u(x, t_j)
+ \sum_{r=1}^K b_{r}\mathcal{M}\big[u_N(x, t_j+c_r\Delta{t})\big].
\end{eqnarray}
%
%
With the starting point $u_N(t_0, x;\Theta_{0})\coloneqq
u_N(t_0, x)$ defined by the initial condition at $t_{0}$, we define
the target function as the sum of squared errors
\begin{eqnarray}
\fl {\rm SSE}_j  = \sum_{s=1}^{K} \Big\| u_N(x, t_j+c_s\Delta{t}; \Theta_{j+1})
  - u_N(x, t_j; \Theta_{j}) - \sum_{r=1}^K a_{sr}
  \mathcal{M}[u_N(x, t_j + c_r\Delta{t}; \Theta_{j+1})]\Big\|_{2}^{2} \nonumber \\
\fl  \hspace{2.2cm} + \Big\| {u}_N(x, t_j+\Delta{t}; \Theta_{j+1}) 
- {u}_N(x, t_j;\Theta_j)
  - \sum_{r=1}^K b_{r}\mathcal{M}[{u}_N(x, t_j+c_r\Delta{t}; 
\Theta_{j+1})]\Big\|_{2}^{2},
  \label{optimization_goal}
\end{eqnarray}
where the $L^2$ norm is taken over the spatial variable $x$.
Minimization of Eq.~\ref{optimization_goal} provides a numerical
solution at $t_{j+1}$ given its value at $t_j$. If coefficients in the
PDE are sufficiently smooth, we can use the basis function expansion
in Eq.~\ref{numti} for ${u}_{N}$ and find that the weights at the
intermediate Runge--Kutta timesteps can be written as the Taylor
expansion

\begin{equation}
    w_i(t_{j}+c_r\Delta{t};\Theta_j) 
= \sum_{\ell=0}^{\infty} \frac{w^{(\ell)}_{i}(t_{j})}{\ell!}(c_r\Delta{t})^{\ell},
\end{equation}
where ${w}^{(\ell)}_{i}(t_{j})$ is the $\ell^{\text{th}}$ derivative
of $w_i$ with respect to time, evaluated at $t_j$.  Therefore, the
neural network is learning the mapping $t_j+c_s\Delta{t}\rightarrow
\sum_{\ell=0}^{\infty} w^{(\ell)}_{i}(t_{j})
(c_s\Delta{t})^{\ell}/\ell!$ for every $i$ by minimizing the loss
function Eq.~\ref{optimization_goal}.


\begin{example}\hspace{-5pt}{\bf: Solving bounded domain PDEs}
\label{bounded_domain}\\
\rm Before focusing on the application of s-PINNs to PDEs whose solution is
defined in an unbounded domain, we first consider the numerical
solution of a PDE in a bounded domain to compare the performance of
the spectral PINN method (using recently developed adaptive methods)
to that of the standard PINN.

Consider the following PDE:
\begin{eqnarray}
        \partial_{t}u = \left(\frac{x+2}{t+1}\right) \partial_{x}u,\,\,\,  x\in(-1, 1), \nonumber\\
    u(x, 0) = \cos(x + 2),\,\,\,  u(1, t) = \cos (3(t+1)),
\label{bounded}
\end{eqnarray}
which admits the analytical solution $u(x, t) = \cos((t+1)(x+2))$. In
this example, we use Chebyshev polynomials $T_i(x)$ as basis functions
and the corresponding Chebyshev-Gauss-Lobatto quadrature
collocation points and weights such that the boundary $u(1, t)
= \cos (3(t+1))$ can be directly imposed at a
collocation point $x=1$.
 \begin{figure}[htb]
      \begin{center}
      \includegraphics[width=6.2in]{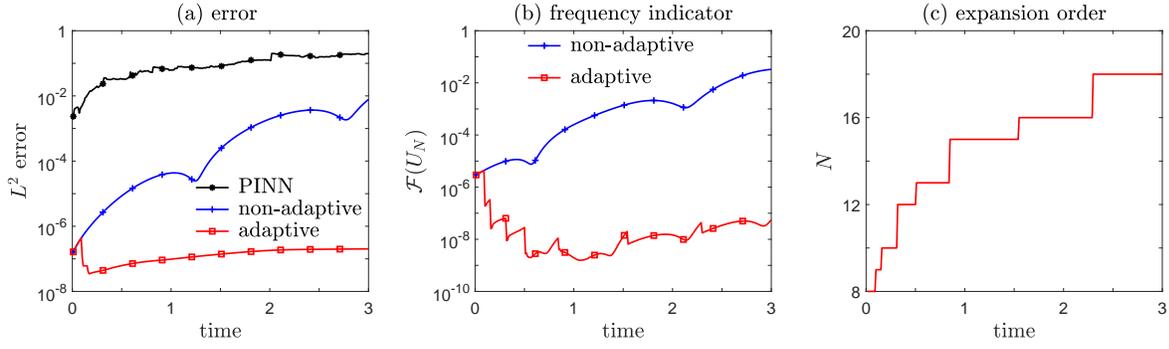}
	\end{center}
      \vspace{-4mm}
      \caption{\small Example~\ref{bounded_domain}: Solving
          Eq.~\ref{bounded} in a bounded domain. $L^2$ errors,
        frequency indicators, and expansion order associated with the
        numerical solution of Eq.~\ref{bounded} using the adaptive
        s-PINN method with a timestep $\Delta{t}=0.01$. (a) In a
        bounded domain, the s-PINNs, with and without the adaptive
        spectral technique, have smaller errors than the standard PINN
        (black). Moreover, the s-PINN method combined with a
        $p$-adaptive technique that dynamically increases the number
        of basis functions (red) exhibits a smaller error than the
        non-adaptive s-PINN (blue).  The higher accuracy of the
        adaptive s-PINN is a consequence of maintaining a small
        frequency indicator~\ref{FREQI}, as shown in (b). (c)
        Keeping the frequency indicator at small values is realized by
        increasing the spectral expansion order.}
     \label{fig_bounded}
\end{figure}

Since the solution becomes increasingly oscillatory in $x$ over time,
an ever-increasing expansion order (\textit{i.e.}, the number of basis
functions) is needed to accurately capture this behavior.  Between
consecutive timesteps, we employ a recently developed $p$-adaptive
technique for tuning the expansion order
\cite{xia2021frequency}. This method is based on monitoring and
controlling a frequency indicator $\mathcal{F}({u}_N)$ defined by
\begin{equation}
    \mathcal{F}(u_N) =
    \left({\f{\sum\limits_{i=N-[\frac{N}{3}]+1}^{N}
        \gamma_{i}(w_{i})^2}{\sum\limits_{i=0}^{N}
\gamma_{i}(w_{i})^2}}\right)^{\f{1}{2}},
    \label{FREQI}
\end{equation}
where $\gamma_{i}\coloneqq \int_{-1}^1 T_{i}^2(x)
(1-x^2)^{-1/2}\dd{x}$. The frequency indicator $\mathcal{F}({u}_N)$
measures the proportion of high-frequency waves and serves as a lower
error bound of the numerical solution ${u}_N(x, t; \Theta)\coloneqq
\sum_{i=0}^N {w}_i(t;\Theta)T_i(x)$. When $\mathcal{F}({u}_N)$ exceeds
its previous value by more than a factor $\rho$, the expansion order
is increased by one.  The indicator is then updated and the factor
$\rho$ also is scaled by a parameter $\gamma \geq 1$.

We use a fourth-order implicit Runge--Kutta method to advance time in
the SSE \ref{optimization_goal} and in order to adjust the expansion
order in a timely way, we take $\Delta{t}=0.01$. The initial expansion
order $N=8$, and the two parameters used to determine the threshold of
adjusting the expansion order are set to $\rho=1.5$ and $\gamma=1.3$.
A neural network with $N_H=4$ layers and $H=200$ neurons per layer is
used in conjunction with the loss function \ref{optimization_goal}
to approximate the solution of Eq.~\ref{bounded}. We compare the
results obtained using the s-PINN method with those obtained using a
fourth-order implicit Runge--Kutta scheme with
$\Delta{x}=\frac{1}{256}, \Delta{t}=0.01$ in a standard PINN approach
\cite{raissi2019physics}, also using $N_H=4$ and $H=200$.

Figure~\ref{fig_bounded} shows that s-PINNs can be used to greatly
improve accuracy because the spectral method can recover exponential
convergence in space, and when combined with a high-order accurate
implicit scheme in time, the overall error is small.  In particular,
the large error shown in Fig.~\ref{fig_bounded} of the standard PINN
suggests that the error of applying auto-differentiation to calculate
the spatial derivative is significantly larger than the spatial
derivatives calculated using spectral methods. Moreover, when
equipping spectral PINNs with the $p$-adaptive technique to
dynamically adjust the expansion order, the frequency indicator can be
controlled, leading to even smaller errors as shown in
Fig.~\ref{fig_bounded}(b,c).

Computationally, using our 4-core laptop on this example, the standard
PINN method requires $\sim 10^{6}$ seconds while the s-PINN approach
with and without adaptive spectral techniques (dynamically increasing
the expansion order $N$) required 1711 and 1008 seconds,
respectively. Thus, s-PINN methods can be computationally more
efficient than the standard PINN approach. This advantage can be
better understood by noting that training of standard PINNs requires
time $\sim {\cal O}(\sum_{i=0}^{N_H}H_i H_{i+1})$ ($H_i$ is the number
of neurons in the $i^{\text{th}}$ layer) to calculate each spatial
derivative (\textit{e.g.}, $\partial_{x}u, \partial_{x}^{2}u,...$) by
autodifferentiation \cite{baydin2018automatic}.  However, in an
s-PINN, since a spectral decomposition ${u}_N(x, t;\Theta)$ has been
imposed, the computational time to calculate derivatives of all orders
is ${\cal O}(N)$, where $N$ is the expansion order.  Since
$\sum_{i=0}^{N_H}H_i H_{i+1}\geq \sum_{i=0}^{N_H} H_i$ and the total
number of neurons $\sum_{i=0}^{N_H} H_i$ is usually much larger than
the expansion order $N$, using s-PINNs can substantially reduce
computational cost.

\end{example}


What distinguishes s-PINNs from the standard PINN framework is that
the latter uses spatial and temporal variables as neural-network
inputs, implicitly assuming that all variables are normalizable
especially when batch-normalization techniques are applied while
training the underlying neural network. However, s-PINNs rely on
spectral expansions to represent the dependence of a function $u(x,t)$
on the spatial variable $x$. Thus, $x$ can be defined in unbounded
domains and does not need to be normalizable. In the following
example, we shall explore how our s-PINN is applied to solving a PDE
defined in $(x, t)\in \mathbb{R}^+\times[0, T]$.


 \begin{figure}[h!]
      \begin{center}
      \includegraphics[width=6.2in]{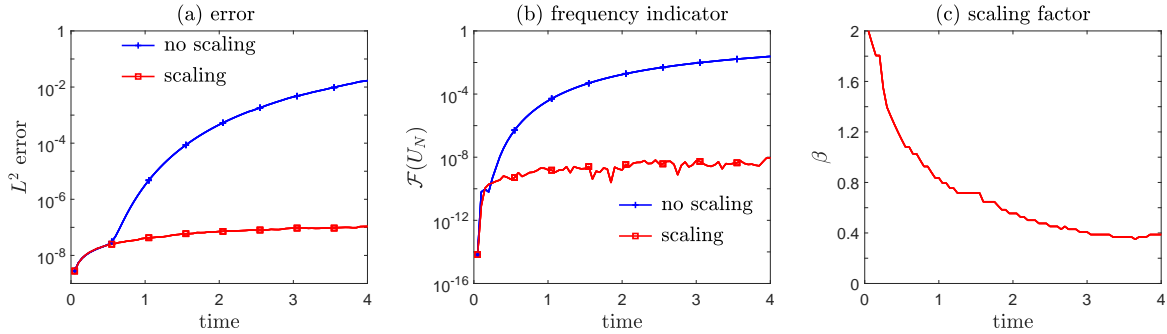}
	\end{center}
      \vspace{-4mm}
      \caption{\small Example~\ref{semi_unbounded_domain}: Solving
        Eq.~\ref{semi_unbounded} in an unbounded domain.  $L^2$ error,
        frequency indicator, and expansion order associated with the
        numerical solution of Eq.~\ref{semi_unbounded} using the
        s-PINN method combined with the spectral scaling
        technique. (a) The s-PINN method with the scaling technique
        (red) has a smaller error than the s-PINN without scaling
        (blue). The higher accuracy of the adaptive s-PINN is a
        consequence of maintaining a smaller frequency indicator
        Eq.~\ref{FREQI}, as shown in (b). (c) Keeping the frequency
        indicator at small values is realized by reducing the scaling
        factor so that the basis functions decay more slowly at
        infinity. The timestep is $\Delta{t}=0.05$}.
     \label{fig_semi_unbounded}
 \end{figure}

\begin{example}\hspace{-5pt}{\bf: Solving unbounded domain PDEs}
\label{semi_unbounded_domain}\\
\rm Consider the following PDE, which is similar to Eq.~\ref{bounded} but
is defined in $(x, t)\in\mathbb{R}^+\times[0, T]$:
\begin{equation}
\partial_{t} u = -\left(\frac{x}{t+1}\right) \partial_{x} u, \quad
u(x, 0) = e^{-x},\,\,\, u(0, t) = 1.
\label{semi_unbounded}
\end{equation}
Equation~\ref{semi_unbounded} admits the analytical solution $u(x, t)
= \exp[-x/(t+1)]$. In this example, we use the basis functions
$\{\hat{\mathcal{L}}_i^{\beta}(x)\}\coloneqq
\{\hat{\mathcal{L}}_i^{(0)}(\beta x)\}$ where
$\hat{\mathcal{L}}_i^{(0)}(x)$ is the generalized Laguerre function of
order $i$ defined in \cite{shen2011spectral}. Here, we use the
Laguerre-Gauss quadrature collocation points and weights so
that $x=0$ is \textit{not} included in the collocation node
set. We use a fourth-order implicit Runge--Kutta method to minimize
the SSE~\ref{optimization_goal} by advancing time. In order to address
the boundary condition, we augment the loss function in
Eq.~\ref{optimization_goal} with terms that represent the cost of
deviating from the boundary condition:
\begin{eqnarray}
\fl  {\rm SSE}_j  = \sum_{s=1}^{K} \Big\| u_N(x, t_j+c_s\Delta{t}; \Theta_{j+1})
  - u_N(x, t_j; \Theta_{j}) - \sum_{r=1}^K a_{sr}
  \mathcal{M}[{u}_N(x, t_j + c_r\Delta{t}; \Theta_{j+1})]\Big\|_{2}^{2} \nonumber  \\[-4pt]
\fl \hspace{2cm} + \Big\| {u}_N(x, t_j+\Delta{t}; \Theta_{j+1}) 
- {u}_N(x, t_j;\Theta_j)
  - \sum_{r=1}^K b_{r}\mathcal{M}[{u}_N(x, t_{j}
+c_r\Delta{t}; \Theta_{j+1})]\Big\|_{2}^{2} \label{optimization_goal_bc} \\[-4pt]
\fl \hspace{1.6cm}  + \sum_{s=1}^K \big[{u}_{N}(0, t_j+c_s\Delta{t}; \Theta_{j+1}) 
- u(0, t_j+c_s\Delta{t})\big]^2 + \big[{u}_{N}(0, t_{j+1}; \Theta_{j+1}) - u(0, t_{j+1})\big]^2, \nonumber
\end{eqnarray}
where the last two terms push the constraints associated with the
Dirichlet boundary condition at $x=0$ at all time points:
\begin{equation}
{u}_{N}(0, t_j+c_s\Delta{t}; \Theta_{j+1}) = u(0,
t_j+c_s\Delta{t}), \quad {u}_{N}(0, t_{j+1}; \Theta_{j+1}) = u(0,t_{j+1}),
\end{equation}
where in this example, $u(0, t_j+c_s\Delta{t})= u(0,t_{j+1}) \equiv
1$.

Because the solution of Eq.~\ref{semi_unbounded} becomes more
diffusive with $x$ (\textit{i.e.}, decays more slowly at infinity), it
is necessary to decrease the scaling factor $\beta$ to allow basis
functions to decay more slowly at infinity.  Between consecutive
timesteps, we adjust the scaling factor by applying the scaling
algorithm proposed in \cite{xia2021efficient}.  Thus, we dynamically
adjust the basis functions in Eq.~\ref{spectral_approx}. As with the
$p$-adaptive technique we used in Example~\ref{bounded_domain}, the
scaling technique also relies on monitoring and controlling the
frequency indicator given in Eq.~\ref{FREQI}. In order to efficiently
and dynamically tune the scaling factor, we set $\Delta{t}=0.05$. The
initial expansion order is $N=8$, the initial scaling factor is
$\beta=2$, the scaling factor adjustment ratio is set to $q=0.95$, and
the threshold for tuning the scaling factor is set to $\nu =
1/(0.95)$. A neural network with 10 layers and 100 neurons per layer
is used in conjunction with the loss function
\ref{optimization_goal}. The neural network of the standard PINN
consists of eight intermediate layers with 200 neurons per layer.
Figure~\ref{fig_semi_unbounded}(a) shows that s-PINNs can achieve very
high accuracy even when a relatively large timestep ($\Delta{t}=0.05$)
is used. Scaling techniques to dynamically control the frequency
indicator are also successfully incorporated into s-PINNs, as shown in
Figs.~\ref{fig_bounded}(b,c).

In Eq.~\ref{semi_unbounded}, we imposed a Dirichlet boundary
condition by modifying the SSE~\ref{optimization_goal_bc} to include
boundary terms. Other types of boundary conditions can be applied in
s-PINNs by including boundary constraints in the SSE as in standard
PINN approaches.

\end{example}


In the next example, we focus on solving a PDE with two spatial
variables, $x$ and $y$, each defined on an unbounded domain.

\begin{example}\hspace{-5pt}{\bf: Solving 2D unbounded domain PDEs}
\label{example_2D}\\
\rm Consider the two-dimensional heat equation on $(x, y)\in\mathbb{R}^2$
\begin{equation}
  \partial_{t}u(x, y, t) = \Delta{u}(x, y, t), \quad
  u(x, y, 0) =  \frac{1}{\sqrt{2}}e^{-x^{2}/12-y^{2}/8}, 
    \label{heatequation}
\end{equation}
which admits the analytical solution
\begin{eqnarray}
\fl  \hspace{2.4cm}u(x, y, t) = \frac{1}{\sqrt{(t+3)(t+2)}}\exp
\left[-\frac{x^2}{4(t+3)}-\frac{y^2}{4(t+2)}\right].
\end{eqnarray}
Note that the solution spreads out over time in both dimensions,
\textit{i.e.}, it decays more slowly at infinity as time increases.
Therefore, we apply the scaling technique to capture the increasing
spread by adjusting the scaling factors $\beta_x$ and $\beta_y$ of the
generalized Hermite basis functions. Generalized Hermite functions of
orders $i=0,\ldots,N_x$ and $\ell=0,\ldots,N_y$ are used in the $x$
and $y$ directions, respectively.

In order to solve Eq.~\ref{heatequation}, we multiply it by any test
function $v\in{H}^1(\mathbb{R})$ and integrate the resulting equation
by parts to convert it to the weak form $(\partial_{t}u, v) =
-(\nabla{u}, \nabla{v})$. Solving the weak form of
Eq.~\ref{heatequation} ensures numerical stability.  When implementing
the spectral method, the goal is to find
  
\begin{equation}
u_{N_x, N_y}^{\beta_x, \beta_y}(x, y, t) 
= \sum_{i=0}^{N_x}\sum_{\ell=0}^{N_y}w_{i, \ell}(t)
\hat{\mathcal{H}}_{i, 0}^{\beta_x}(x)
\hat{\mathcal{H}}_{\ell, 0}^{\beta_y}(y),
\label{tensor_prod}
\end{equation} 
where $\hat{\mathcal{H}}_{i, 0}^{\beta_x},\, \hat{\mathcal{H}}_{\ell,
  0}^{\beta_y}$ are generalized Hermite functions defined in
Table~\ref{tab:model_variables} such that $(\partial_{t}u, v) = -(\nabla{u},
\nabla{v})\,\,t\in(t_j, t_{j+1})$ for all $v\in P_{N_x,
  0}^{\beta_x}\times P_{N_y, 0}^{\beta_y},\, t\in(t_j, t_{j+1})$.
This allows one to advance time from $t_j$ to $t_{j+1}$ given $u_{N_x,
  N_y}^{\beta_x, \beta_y}(x, t_j)$.

\begin{figure}[htb]
      \begin{center}
      \includegraphics[width=4.8in]{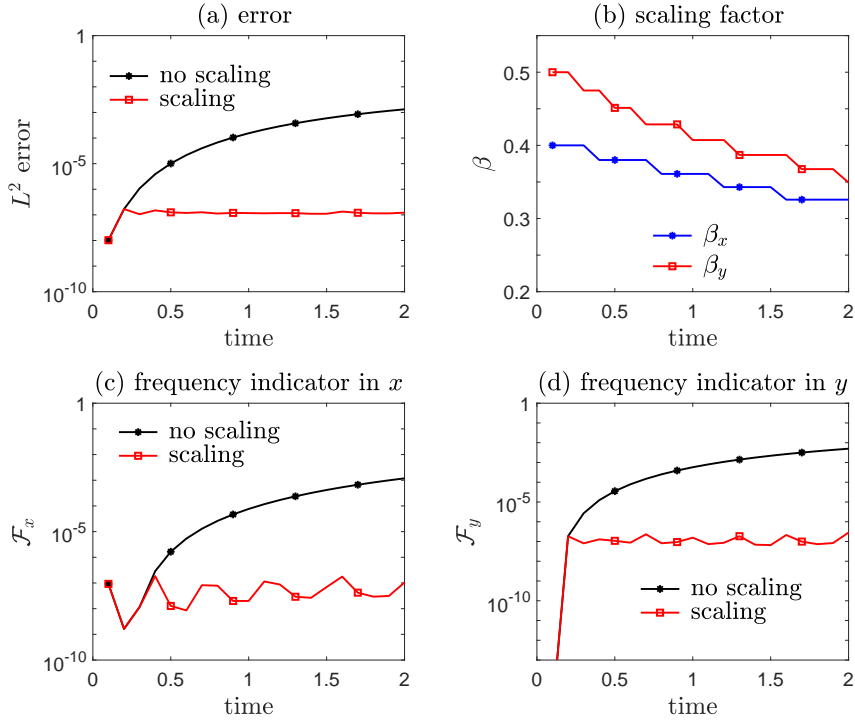}
      \end{center}
      \vspace{-4mm}
       \caption{\small Example~\ref{example_2D}: {Solving a higher
           dimensional unbounded domain PDE
           (Eq.~\ref{heatequation}).}  $L^2$ error, scaling factor,
         and frequency indicators associated with the numerical
         solution of Eq.~\ref{heatequation} using s-PINNs, with and
         without dynamic scaling. (a) $L^2$ error as a function of
         time. The s-PINNs that are equipped with the scaling
         technique (red) achieve higher accuracy than those without
         (black). (b) The scaling factors $\beta_x$ (blue) and
         $\beta_y$ (red) as functions of time. Both scaling factors
         are decreased to match the spread of the solution in both the
         $x$ and $y$ directions. Scaling factors are adjusted to
         maintain small frequency indicators in the $x$-direction (c),
         and in the $y$-direction (d). In all computations,
         the timestep is $\Delta{t}=0.1$.}
     \label{fig6}
\end{figure}

Tuning the scaling factors $\beta_x, \beta_y$ across different
timesteps is achieved by monitoring the frequency indicators in the
$x$- and $y$-directions, $\mathcal{F}_x$ and $\mathcal{F}_y$, as
detailed in \cite{xia2021efficient}. We use initial expansion orders
$N_x=N_y=8$ and scaling factors $\beta_x=0.4, \beta_y=0.5$. The ratio
and threshold for adjusting the scaling factors, are set to be
$q=0.95$ and $\nu^{-1}=0.95$.  The timestep $\Delta{t}=0.1$ is used to
adjust both scaling factors in both dimensions in a timely manner and a
fourth order implicit Runge--Kutta scheme is used for numerical
integration. The neural network that we use to learn ${w}_{i,
  \ell}(t)$ has 5 intermediate layers with 150 neurons in each layer.

The results depicted in Fig.~\ref{fig6}(a) show that an s-PINN using
the scaling technique can achieve high accuracy by using high-order
Runge--Kutta schemes in minimizing the SSE~\ref{optimization_goal}
and by properly adjusting $\beta_x$ and $\beta_y$ (shown in
Fig.~\ref{fig6}(b)) to control the frequency indicators
$\mathcal{F}_x$ and $\mathcal{F}_y$ (shown in Fig.~\ref{fig6}(c) and
(d)). The s-PINNs can be extended to higher spatial dimensions by
calculating the numerical solution expressed in tensor product form as
in Eq.~\ref{tensor_prod}. 

\end{example}

Since our method outputs spectral expansion coefficients, if using the
full tensor product in the spatial spectral decomposition leads to a
number of outputs that increase exponentially with dimensionality.
The very wide neural networks needed for such high-dimensional
problems results in less efficient training.  However, unlike other
recent machine\textendash learning\textendash based PDE solvers or PDE
learning methods \cite{brandstetter2021message,li2020fourier} that
explicitly rely on a spatial discretization of grids or meshes, the
curse of dimensionality can be partially mitigated in our s-PINN
method. By using a hyperbolic cross space \cite{shen2010sparse}, we
can effectively reduce the number of coefficients needed to accurately
reconstruct the numerical solution. In the next example, we solve a 3D
parabolic spatiotemporal PDE, similar to that in
Example~\ref{example_2D}, but we demonstrate how implementing a
hyperbolic cross space can reduce the number of outputs and boost
training efficiency.

\begin{example}\hspace{-5pt}{\bf: Solving 3D unbounded domain PDEs}
\label{example_3D}\\
\rm 
Consider the (3+1)-dimensional heat equation
\begin{equation}
  \partial_{t}u(x, y, z, t) = \Delta{u}(x, y, z, t), \quad
  u(x, y, 0) =  \frac{1}{\sqrt{6}}e^{-x^{2}/12-y^{2}/8-z^2/4}, 
    \label{heatequation3D}
\end{equation}
which admits the analytical solution
\begin{eqnarray}
\fl \hspace{1.2cm} u(x, y, z, t) = \frac{1}{\sqrt{(t+3)(t+2)(t+1)}}\exp
\left[-\frac{x^2}{4(t+3)}-\frac{y^2}{4(t+2)} - \frac{z^2}{4(t+1)}\right]
\end{eqnarray}
for $(x, y, z)\in\mathbb{R}^3$. If we use the full tensor product of
spectral expansions with expansion orders $N_x=N_y=N_z=9$, we will
need to output $10^3=1000$ expansion coefficients, and in turn, a
relatively wide neural network with many parameters will be needed to
generate the corresponding weights as shown in
Fig.~\ref{fig:spectral_pinn}(c). Training such wide networks can be
inefficient. However, many of the spectral expansion coefficients are
close to zero and can be eliminated without compromising accuracy. One
way to select expansion coefficients is to use the hyperbolic cross
space technique~\cite{shen2010sparse} to output coefficients of the
generalized Hermite basis functions only in the space

\begin{eqnarray}
  & V_{N, \gamma_{\times}}^{\vec{\beta}, \vec{x}_0} \coloneqq \text{span}
  \Big\{\hat{\mathcal{H}}_{n_1}(\beta^{1}x)
  \hat{\mathcal{H}}_{n_2}(\beta^2y)\hat{\mathcal{H}}_{n_3}(\beta^{3}z):
  |\vec{n}|_{\text{mix}}\|
  \vec{n}\|_{\infty}^{-\gamma_{\times}}\leq N^{1-\gamma_{\times}} \Big\}, \nonumber \\ &
  \vec{n}\coloneqq(n_1, n_2, n_3),\,\,  ~|\vec{n}|_{\text{mix}} \coloneqq
  \max\{n_1, 1\} \max\{n_2, 1\} \max\{n_3, 1\},
\label{hyper_space} 
\end{eqnarray}
where the hyperbolic space index $\gamma_{\times}\in (-\infty, 1)$.
Taking $\gamma_{\times}=-\infty$ in Eq.~\ref{hyper_space} corresponds
to the full tensor product with $N+1$ basis functions in each
dimension. For fixed $N$ in Eqs.~\ref{hyper_space}, the number of
total basis function tend to decrease with increasing
$\gamma_{\times}$. We set $N=9$ in Eq.~\ref{hyper_space} and use the
initial scaling factors $\beta_x=0.4, \beta_y=0.5, \beta_z=0.7$. Using
a fourth-order implicit Runge--Kutta scheme with timestep
$\Delta{t}=0.2$, we set the ratio and threshold for adjusting the
scaling factors are set to $q=0.95$ and $\nu^{-1}=0.95$ in each
dimension.

To illustrate the potential numerical difficulties arising from
outputting large numbers of coefficients when solving
higher-dimensional spatiotemporal PDEs, we use a neural network with
two hidden layers and different numbers of neurons in the intermediate
layers. We also adjust $\gamma_{\times}$ to explore how decreasing the
number of coefficients can improve training efficiency. Our results
are listed Table \ref{tab:hyper_cross}.

\begin{table}
  \centering
  \caption{{\small Example~\ref{example_3D}: Applying hyperbolic cross
      space and s-PINNs to the (3+1) dimensional PDE
      Eq.~\ref{heatequation3D}. Applying the hyperbolic cross space
      (Eq.~\ref{hyper_space}), we record the $L^2$ error in the lower
      left and the training time in the upper right of each cell. The
      number of coefficients (outputs in the neural network) for
      $\gamma_{\times}=-\infty, -1, 0, \frac{1}{2}$ are $1000, 205,
      141, 110$, respectively. Using $\gamma_{\times}=-1$ or $0$ leads
      to the most accurate results. The training time tends to
      increase with the number of outputs (a smaller $\gamma_{\times}$
      corresponds to more outputs). By comparing the results in
      different rows for the same column, it can be seen that more
      outputs require a wide neural network for training.}}
  \label{tab:hyper_cross}
\vspace{0.1in}
\scriptsize
\renewcommand*{\arraystretch}{1.2}
    \begin{tabular}{|l|r|r|r|r|r|}
\hline
\diagbox{$H$}{$\gamma_{\times}$} & $-\infty$ & $-1$ & 0 & $\frac{1}{2}$ \\[2pt]
\hline
    200 &  \diagbox{2.217e-03}{22911}&  \diagbox{1.651e-04}{4309}&\diagbox{5.356e-05}{2886}& \diagbox{3.173e-04}{3956}\\[2pt]
\hline
    400 &  \diagbox{1.072e-03}{26725}&  \diagbox{2.970e-05}{7014}&\diagbox{5.356e-05}{3309}& \diagbox{3.173e-04}{2356}\\[2pt]
\hline
    700  &  \diagbox{2.276e-03}{43923}& \diagbox{2.900e-05}{3133}& \diagbox{5.356e-05}{3229}&\diagbox{3.173e-04}{2098}\\[2pt]
\hline
1000  &  \diagbox{7.871e-05}{55880} &
\diagbox{2.901e-05}{3002} & \diagbox{5.356e-05}{2016} & 
\diagbox{3.173e-04}{1894}\\[2pt]
\hline
    \end{tabular}%
\end{table}

The results shown in Table~\ref{tab:hyper_cross} indicate that,
compared to using the full tensor product $\gamma_{\times}=-\infty$,
implementing the hyperbolic cross space with a moderate
$\gamma_{\times}=-1$ or $0$, the total number of outputs is
significantly reduced, leading to faster training and better
accuracy. However, increasing the hyperbolicity to
$\gamma_{\times}=\frac{1}{2}$, the error increases relative to using
$\gamma_{\times}=-1, 0$ because some useful, nonzero coefficients are
excluded. Also, comparing the results across different rows, wider
layers lead to both more accurate results and faster training
speed. The sensitivity of our s-PINN method to the number of
intermediate layers in the neural network and the number of neurons in
each layer are further discussed in
Example~\ref{example_sensitivity}. Overall, in higher-dimensional
problems, there is a balance between computational cost and accuracy
as the number of outputs needed will grow fast with dimensionality.
Spectrally-adapted PINNs can easily incorporate a hyperbolic cross
space so that the total number of outputs can be reduced to a
manageable number for moderate-dimensional problems. Finding the
optimal hyperbolicity index $\gamma_{\times}$ for the cross space
Eq.~\ref{hyper_space} will be problem-specific.
%
%

\end{example}

In the next example, we explore how s-PINNs can be used to solve
Schr\"odinger's equation in $x\in\mathbb{R}$. Solving this
complex-valued equation poses substantial numerical difficulties as
the solution exhibits diffusive, oscillatory, and convective
behavior~\cite{li2018stability}.


\begin{example}\hspace{-5pt}{\bf: Solving an unbounded domain Schr\"{o}dinger equation}
\label{schrodinger}\\
\rm We seek to numerically solve the following Schr\"odinger equation
defined on $x\in \mathbb{R}$
\begin{equation}
  \text{i}\partial_t \psi(x, t) = -\partial_x^2\psi(x, t), \quad
  \psi(x, 0) = \frac{1}{\sqrt{\zeta}}\exp\left[\text{i}kx-
\frac{x^2}{4\zeta}\right].
\label{Shrodinger2}
\end{equation}
For reference, Eq.~\ref{Shrodinger2} admits the analytical solution
\begin{equation}
    \psi(x, t) = \frac{1}{\sqrt{\zeta+\text{i}t}}
\exp\left[\text{i}k(x-kt)-\frac{(x-2kt)^2}{4(\zeta+\text{i}t)}\right].
\end{equation}
As in Example~\ref{example_2D}, we shall numerically solve
Eq.~\ref{Shrodinger2} in the weak form
\begin{equation}
(\partial_t \Psi(x, t), v) + \text{i}(\partial_x \Psi(x, t), \partial_x v)
=0, \quad \forall v\in H^1(\mathbb{R}).
\label{numweak2}
\end{equation} 
Since the solution to Eq.~\ref{Shrodinger2} decays as $\sim
\exp[-{x^2}/{(4\sqrt{(\zeta^2+t^2)})}]$ at infinity, we shall use the
generalized Hermite functions as basis functions. The solution is
rightward-translating for $k>0$ and increasingly oscillatory and
spread out over time. Hence, as detailed in \cite{xia2021frequency},
we apply three additional adaptive spectral techniques to improve
efficiency and accuracy: (i) a scaling technique to adjust the scaling
factor $\beta$ over time in order to capture diffusive behavior, (ii)
a moving technique to adjust the center of the basis function $x_L$ to
capture convective behavior, and (iii) a $p$-adaptive technique to
increase the number of basis functions $N$ to better capture the
oscillations. We set the initial parameters $\beta=0.8, x_L=0, N=24$
at $t=0$. The scaling factor adjustment ratio and the threshold for
adjusting the scaling factor are $q=\nu^{-1}=0.95$, the minimum and
maximum displacements are $0.004$ and $0.1$ within each timestep for
moving the basis functions, respectively, and the threshold for moving
is $1.001$. Finally, the thresholds of the $p$-adaptive technique are
set to $\rho=\rho_0=2$ and $\gamma=1.4$. To numerically solve
Eq.~\ref{numweak2}, a fourth-order implicit Runge--Kutta scheme is
applied to advance time with timestep $\Delta{t}=0.1$. The neural
network underlying the s-PINN that we use in this example contains 13
layers with 80 neurons in each layer.
 \begin{figure}[htb]
      \begin{center}
      \includegraphics[width=4.8in]{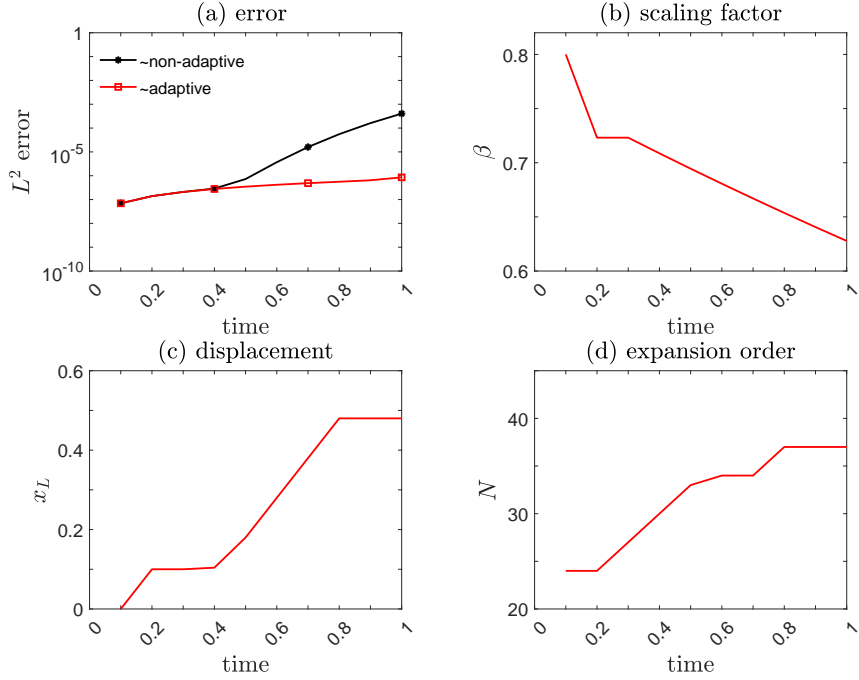}
	\end{center}
      \vspace{-4mm}
        \caption{\small Example~\ref{schrodinger}: {Solving the
            Schr\"{o}dinger equation (Eq.~\ref{Shrodinger2}) in an
            unbounded domain.} Approximation error, scaling factor,
          displacement, and expansion order associated with the
          numerical solution of Eq.~\ref{Shrodinger2} using adaptive
          (red) and non-adaptive (black) s-PINNs. (a) Errors for
          numerically solving Eq.~\ref{Shrodinger2} with and without
          adaptive techniques. (b) The change of the scaling factor
          which decreases over time as the solution becomes more
          spread out. (c) The displacement of the basis functions
          $x_L$ which is increased as the solution moves
          rightwards. (d) The expansion order $N$ increases over time
          as the solution becomes more oscillatory. A timestep
          $\Delta{t}=0.1$ was used.}
     \label{fig_schrodinger}
\end{figure}
Figure~\ref{fig_schrodinger}(a) shows that the s-PINN with adaptive
spectral techniques leads to very high accuracy as it can
properly adjust the basis functions over a longer timescale (across
different timesteps), while not adapting the basis functions results
in larger errors. Figs.~\ref{fig_schrodinger}(b--d) show that the
scaling factor $\beta$ decreases over time to match the spread of the
solution, the displacement of the basis function $x_L$ increases in
time to capture the rightward movement of the basis functions, and the
expansion order $N$ increases to capture the solution's increasing
oscillatory behavior. Our results indicate that our s-PINN method can
effectively utilize all three adaptive algorithms.

\end{example}


We now explore how the timestep and the order of the implicit
Runge--Kutta method affect the approximation error, \textit{i.e.}, to
what extent can we relax the constraint on the timestep and maintain
the accuracy of the basis functions, or, if higher-order Runge--Kutta
schemes are better. Another feature to explore is the neural network
structure, such as the number of layers and neurons per layer, and how
it affects the performance of s-PINNs. In the following example, we
carry out a sensitivity analysis.


\begin{example}\hspace{-5pt}{\bf: Sensitivity analysis of s-PINN}
\label{example_sensitivity}\\
\rm To explore how the performance of an s-PINN depends on algorithmic
set-up and parameters, we apply it to solving the heat equation
defined on $x\in \mathbb{R}$,
\begin{equation}
\partial_t u(x,t)  =  \partial_{x}^{2} u(x,t) + f(x,t),\quad  u(x, 0)  =  e^{-x^{2}/4} \sin x
\label{parabolic}
\end{equation}
using generalized Hermite functions as basis functions.  For the
source $f(x, t) = [x\cos x + (t+1)\sin x] \, (t+1)^{-3/2} \,
\exp[-\frac{x^2}{4(t+1)}]$, Eq.~\ref{parabolic} admits the
analytical solution
\begin{equation}
    u(x, t) = \frac{\sin x}{\sqrt{t+1}} \exp\left[-\frac{x^2}{4(t+1)}\right].
\label{soln6}
\end{equation}
We solve Eq.~\ref{parabolic} in the weak form by multiplying
any test function $v\in{H}^1(\mathbb{R})$ on both sides and
integrating by parts to obtain
\begin{equation}
    (\partial_{t}u, v) = -(\partial_{x}u, \partial_{x}v) 
+ (f, v),\,\,\, \forall v\in H^1(\mathbb{R}).
    \label{para_weak}
\end{equation}
The solution diffusively spreads over time, requiring one to decrease
the scaling factor $\beta$ of the generalized Hermite functions
$\{\hat{\mathcal{H}}^{\beta}_i(x)\}$. We shall first study how the
timestep and the order of the implicit Runge--Kutta method associated
with solving the minimization problem \ref{optimization_goal} affect
our results. We use a neural network with five intermediate layers and
200 neurons per layer, and set the learning rate
$\eta=5\times10^{-4}$. The initial scaling factor is set to
$\beta=0.8$. The scaling factor adjustment ratio and threshold are set
to $q=0.98$, and $\nu=q^{-1}$, respectively.  For comparison, we also
apply a Crank-Nicolson scheme for numerically solving
Eq.~\ref{para_weak}, \textit{i.e.},


\begin{equation}
\frac{U_N^{\beta}(t_{j+1}) - U_N^{\beta}(t_{j})}{\Delta t} = 
D_N^{\beta}\frac{\big[U_N^{\beta}(t_{j+1}) + U_N^{\beta}(t_{j})\big]}{2}
+ \frac{F_N^{\beta}(t_{j+1})+F_N^{\beta}(t_{j})}{2}.
\label{CKscheme}
\end{equation}
where $U_N^{\beta}(t), F_N^{\beta}(t)$ are the $N+1$-dimensional
vectors of spectral expansion coefficients of the numerical solution
and the of the source, respectively.
$D_N^{\beta}\in\mathbb{R}^{(N+1)\times(N+1)}$ is the tridiagonal block
matrix representing the discretized Laplacian operator $\partial_x^2$:

\begin{equation*}
    D_{i, i-2} = \beta^2\frac{\sqrt{(i-2)(i-1)}}{2}, \,\,\,
D_{i,i} = -\beta^2\Big( i-\frac{1}{2}\Big), \,\,\,
D_{i,i+2} =\beta^2\frac{\sqrt{i(i+1)}}{2},
\end{equation*}
and $D_{i,j}=0$, otherwise.


\begin{table}
  \centering
  \caption{{\small Example~\ref{example_sensitivity}: Sensitivity
      analysis of s-PINN. Computational runtime (in seconds), error,
      and the final scaling factor for different timesteps $\Delta t$,
      different implicit order-$K$ Runge--Kutta schemes, and the
      traditional Crank-Nicolson scheme. In each box, the run time (in
      seconds), the SSE, and the final scaling factor are listed from
      left to right. The results associated with the smallest error
      are highlighted in red while the results associated with the
      shortest run time for our s-PINN method are indicated in blue.}}
  \label{tab:qdtime}
\vspace{3mm}
 \scriptsize
\renewcommand*{\arraystretch}{1.2}
    \begin{tabular}{|l|r|r|r|r|r|}
\hline
\diagbox{$\Delta{t}$}{$K$} & C-K scheme & 2 & 4 & 6 & 10 \\
\hline
    0.02 &  12, 8.252e-06, 0.545 & 27, 4.011e-08, 0.545 & \textcolor{red}{54, 1.368e-08, 0.545} & 
279, 2.545e-07, 0.545 & 7071, 6.358e-05, 0.695\\
\hline
   0.05 &5, 5.157e-05, 0.545 & 12, 2.799e-08, 0.545 & 23, 1.651e-08, 0.545 & 
105, 2.566e-07, 0.545 & 3172, 1.052e-06, 0.545\\
\hline
    0.1 & 3, 2.239e-04, 0.695 & 6, 1.331e-06, 0.695  & 10, 1.314e-06, 0.695 & 72, 1.346e-06, 0.695 &
1788, 2.782e-06, 0.695 \\
\hline
    0.2 & 2, 9.308e-04, 0.695 &\textcolor{blue}{3, 3.760e-06, 0.695}  & 9, 2.087e-06, 0.695 & 
317, 2.107e-06, 0.695 & 1310, 1.925e-03, 0.753 \\
\hline
    \end{tabular}%
\end{table}

Table~\ref{tab:qdtime} shows that since the error from temporal
discretization $\Delta{t}^{2K}$ is already quite small for $K \geq 4$,
using a higher-order Runge--Kutta method does not significantly
improve accuracy for all choices of $\Delta{t}$. Using higher-order
($K\geq 4$) schemes tends to require longer run times. Higher orders
require fitting over more data points (using the same number of
parameters) leading to slower convergence when minimizing
Eq.~\ref{optimization_goal}, which can result in larger
errors. Compared to the second-order Crank-Nicolson scheme, whose
error is $~O(\Delta{t}^2)$, the errors of our s-PINN method do not
grow significantly when $\Delta{t}$ increases. In fact, the accuracy
using the smallest timestep $\Delta{t}=0.02$ in the Crank-Nicolson
scheme was still inferior to that of the s-PINN method using the
second order or fourth order Runge-Kutta scheme with $\Delta{t}=0.2$.
Moreover, the run time of our s-PINN method using a second or
fourth-order implicit Runge--Kutta scheme for the loss function is not
significantly larger than that of the Crank-Nicolson scheme. Thus,
compared to traditional spectral methods for numerically solving PDEs,
our s-PINN method, even when incorporating some lower-order
Runge--Kutta schemes, can greatly improve accuracy without
significantly increasing computational cost.

In Table~\ref{tab:qdtime}, the smallest run time of our s-PINN method,
which occurs for $K=2, \Delta{t}=0.2$, is shown in blue. The smallest
error case, which arises for $K=4, \Delta{t}=0.02$, is shown in
red. The run time always increases with the order $K$ of the implicit
Runge--Kutta scheme and always decreases with $\Delta{t}$ due to fewer
timesteps.  Additionally, the error always increases with $\Delta{t}$
regardless of the order of the Runge--Kutta scheme. However, the
expected convergence order is not observed, implying that the increase
in error results from increased lag in adjustment of the scaling
factor $\beta$ when $\Delta{t}$ is too large, rather than from an
insufficiently small time discretization error $\Delta{t}^{2K}$.
Using a fourth-order implicit Runge--Kutta scheme with
$\Delta{t}=0.05$ to solve Eq.~\ref{para_weak} seems to both achieve
high accuracy and avoid large computational costs.

We also investigate how the total number of parameters in the neural
network and the structure of the network affect efficiency and
accuracy. We use a sixth-order implicit Runge--Kutta scheme with
$\Delta{t}=0.1$. The learning rate is set to $\eta=5\times 10^{-4}$
for all neural networks.

\begin{table}
  \centering
  \caption{{\small Example~\ref{example_sensitivity}: Sensitivity
      analysis of s-PINN.  Computational runtime (in seconds), error,
      and the final scaling factor for different numbers of
      intermediate layers $N_H$ and neurons per layer $H$. In each
      box, the run time (in seconds), the SSE, and the final scaling
      factor are listed from left to right. Results associated with
      the smallest error are marked in red while those associated with
      the shortest run time are highlighted in blue.}}
  \label{tab:hnh}
\vspace{3mm}
\scriptsize
\renewcommand*{\arraystretch}{1.2}
    \begin{tabular}{|l|r|r|r|r|r|}
\hline
\diagbox{$H$}{$N_H$} & 3 & 5 & 8 & 13 \\[2pt]
\hline
    50 &  1348, 6.317e-04, 0.738 &  798, 9.984e-05, 0.695 & 995, 1.891e-04, 0.579 & 778, 4.022e-04, 0.695\\[2pt]
\hline
    80 &  784, 7.164e-04, 0.654 &  234, 1.349e-06, 0.695 & 216, 1.345e-06, 0.695 & 376, 1.982e-06, 0.695\\[2pt]
\hline
    100  &  1080, 8.804e-05, 0.695 & 114, 1.344e-06, 0.695 & 102, 1.346e-06, 0.695 & 145, 1.348e-06, 0.695\\[2pt]
\hline
    200  &  219, 1.349e-06, 0.695 & 72, 1.346e-06, 0.695 & 43, 1.347e-06, 0.695 & 
64, 1.345e-06, 0.695\\[2pt]
\hline
    \end{tabular}%
\end{table}
As shown in Table~\ref{tab:hnh}, the computational cost tends to
decrease with the number of neurons $H$ in each layer as it takes
fewer epochs to converge when minimizing Eq.~\ref{optimization_goal}.
The run time tends to decrease with $N_H$ due to a faster convergence
rate, until about $N_H=8$.  The errors when $H=50$ are significantly
larger as the training terminates (after a maximum of 100000 epochs)
before it converges. For $N_H=3$, the corresponding s-PINN always
fails to achieve accuracy within 100000 epochs unless $H \gtrsim
200$. Therefore, overparametrization is indeed helpful in improving
the neural network's performance, leading to faster convergence rates,
in contrast to most traditional optimization methods that take longer
to converge with more parameters. Similar observations have been made
in other optimization tasks that involve deep neural networks
\cite{pmlr-v80-arora18a,chen2020much}. Consequently, our s-PINN method
retains the advantages of deep and wide neural networks for improving
accuracy and efficiency.
\end{example}
%

\section{Parameter Inference and Source Reconstruction}
\label{model_reconstruction}
As with standard PINN approaches, s-PINNs can also be used for
parameter inference in PDE models or reconstructing unknown sources in
a physical model.  Assuming observational data at uniform time
intervals $t_{j}= j\Delta t$ associated with a partially known
underlying PDE model, s-PINNs can be trained to infer model parameters
$\theta$ by minimizing the sum of squared errors, weighted from both
ends of the time interval $(t_{j}, t_{j+1})$,
\begin{equation}
    {\rm SSE}_{j} = {\rm SSE}_{j}^{\rm L} + {\rm SSE}_{j}^{\rm R},
    \label{MSEgoal}
\end{equation}
where 
\begin{eqnarray}
{\rm SSE}_{j}^{\rm L}  = & \sum_{s=1}^K \Big\| {u}(x, t_{j}+c_s\Delta{t};\theta_{j+1}; {\Theta}_{j+1}) 
- u(x, t_{j}; \theta_{j}) \nonumber \\[-8pt]
\: & \hspace{2.5cm}  - \sum_{r=1}^K 
a_{sr}\mathcal{M}\big[{u}(x, t_{j}+c_r\Delta{t};\theta_{j+1};{\Theta}_{j+1})\big]
\Big\|_{2}^{2}, \nonumber \\
{\rm SSE}_{j}^{\rm R} = & \sum_{s=1}^K \Big\| {u}(x, t_{j}+c_s\Delta{t};\theta_{j+1};{\Theta}_{j+1}) 
- u(x, t_{j+1}; \theta_{j+1}) \nonumber \\[-8pt]
\: & \hspace{2.3cm} - \sum_{r=1}^K (a_{sr}-b_r)
\mathcal{M}\big[{u}(x, t_{j}+c_r\Delta{t};\theta_{j+1};{\Theta}_{j+1})\big]
\Big\|_{2}^{2}.
\label{MSEdef}
  \end{eqnarray}
Here, $\theta_{j+1}$ is the model parameter to be found using the
sample points $c_{s}\Delta t$ between $t_{j}$ and $t_{j+1}$.  The most
obvious advantage of s-PINNs over standard PINN methods is that they
can deal with models defined on unbounded domains, extending
PINN-based methods that are typically applied to finite domains.

Given observations over a certain time interval, one may wish to both
infer parameters ${\theta}_{j}$ in the underlying physical model
and reconstruct the solution ${u}$ at any given time.  Here, we
provide an example in which both a parameter in the model is to be
inferred and the numerical solution obtained.
%
\begin{example}\hspace{-5pt}{\bf: Parameter (diffusivity) inference}
\label{KAPPA}\\
\rm As a starting point for a parameter-inference problem, we consider
 diffusion with a source defined on  $x\in \mathbb{R}$
\begin{equation}
  \partial_t u(x,t)  = \kappa \partial_{x}^{2} u(x,t)
  + f(x,t), \quad u(x, 0)  = e^{-x^{2}/4}\sin x,
    \label{para_reconstruct}
\end{equation}
where the constant parameter $\kappa$ is the thermal conductivity (or
diffusion coefficient) in the entire domain. In this example, we set
$\kappa=2$ as a reference and assume the source
\begin{equation}
\fl \hspace{7mm} f(x, t) = \left[ {2\left(x \cos x + (t+1) \sin x\right) \over 
 (t+1)^{3/2}} - \frac{x^2}{4(t+1)^{2}} +  \frac{\sin x }{2(t+1)^{3/2}}\right]
\exp\left[-\frac{x^2}{4(t+1)}\right].
\end{equation}
In this case, the analytical solution to Eq.~\ref{para_reconstruct}
is given by Eq.~\ref{soln6}.
%
%
We numerically solve Eq.~\ref{para_reconstruct} in the weak form of
Eq.~\ref{para_weak}. If the form of the spatiotemporal heat equation
is known (such as Eq.~ \ref{para_reconstruct}), but some parameters
such as $\kappa$ is unknown, reconstructing it from measurements is
usually performed by defining and minimizing a loss function as was
done in \cite{huntul2021identification}. It can also be shown that
$\kappa=\kappa(t)$ in Eq.~\ref{para_reconstruct} can be uniquely
determined by the observed solution $u(x, t)$
\cite{ivanchov1993inverse,jones1962determination,beznoshchenko1974on}
under certain conditions. Here, however, we assume that observations
are taken at discrete time points $t_j = j\Delta{t}$ and seek to
reconstruct both the parameter $\kappa$ and the numerical solution at
$t_j+c_s\Delta{t}$ (defined in Eqs.~\ref{MSEdef}) by minimizing
Eq.~\ref{MSEgoal}. We use a neural network with 13 layers and 100
neurons per layer with a sixth-order implicit Runge--Kutta scheme. The
timestep $\Delta{t}$ is 0.1. At each timestep, we draw the function
values from
\begin{equation}
u(x, t_j) = \frac{\sin x }{\sqrt{t_j+1}}
\exp\left[-\frac{x^2}{4(t_j+1)}\right] + \xi(x, t_j),
\label{noisy_data}
\end{equation}
where $\xi(x, t)$ is the noise term that is both spatially and
temporally uncorrelated, and $\xi(x, t)\sim\mathcal{N}(0, \sigma^2)$,
where $\mathcal{N}(0, \sigma^2)$ is the normal distribution of mean 0
and variance $\sigma^2$ (\textit{i.e.}, $\langle \xi(x, t)\xi(t,
s))\rangle=\sigma^2\delta_{x, y}\delta_{s, t}$). For different levels
of noise $\sigma^2$, we take one trajectory of the measured solution
with noise $u(x, t_j)$ to reconstruct the parameter $\kappa$, which is
presumed to be a constant in $[t_j, t_{j+1})$, and simultaneously
  obtain the numerical solutions at the intermediate time points
  $t_j+c_s\Delta{t}$. We are interested in how different levels of
  noise and the increasing spread of the solution will affect the SSE
  and the reconstructed parameter
  $\hat{\kappa}$. Figure~\ref{fig6_lambda} shows the deviation of the
  reconstructed $\hat{\kappa}$ from its true value,
  $|\hat{\kappa}-2|$, the SSE, the scaling factor, and the frequency
  indicator as functions of time for different noise levels.
 \begin{figure}[h!]
      \begin{center}
      \includegraphics[width=4.8in]{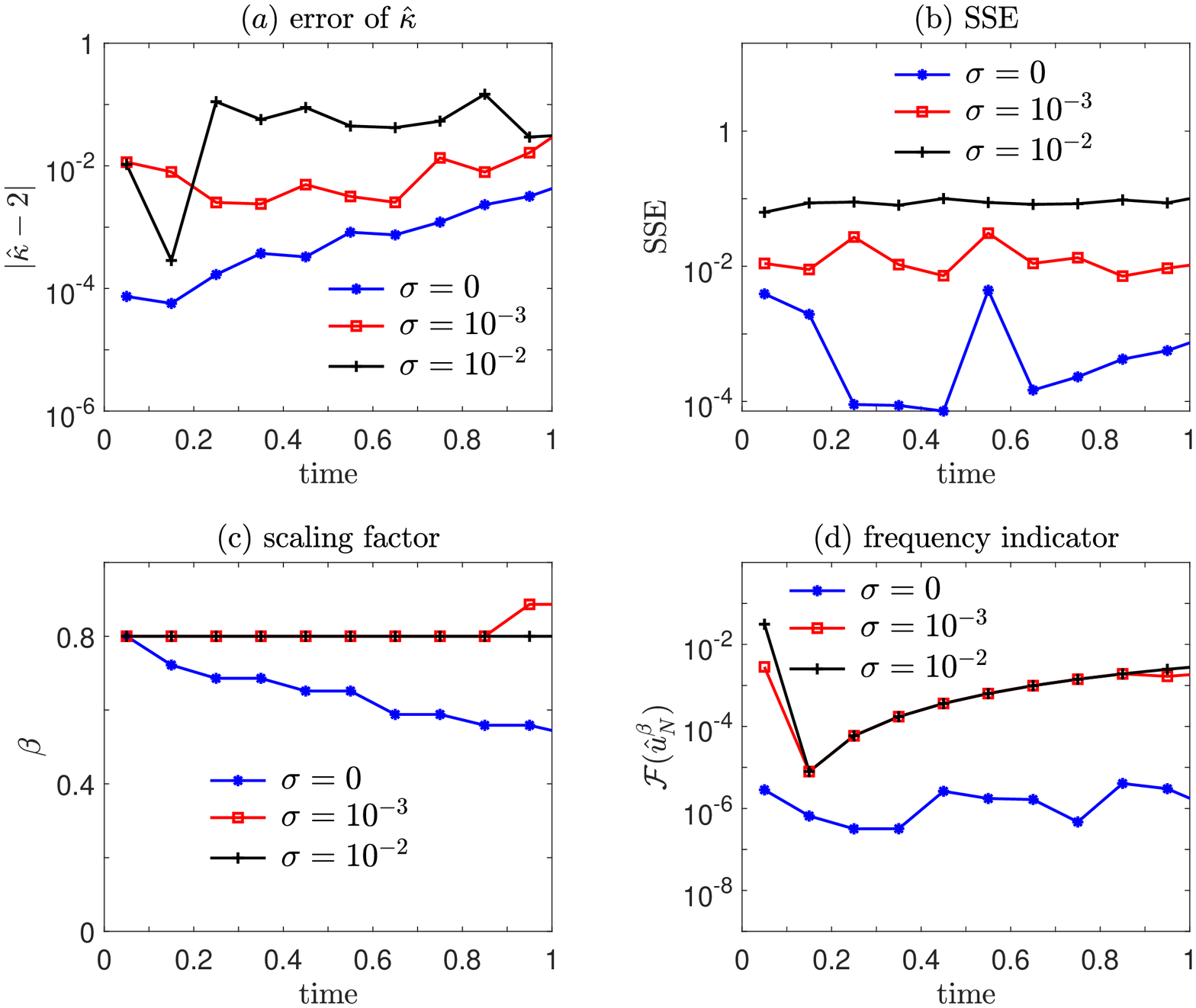}
	\end{center}
      \vspace{-4mm}
        \caption{\small Example~\ref{KAPPA}: Parameter (diffusivity)
          inference.  The parameter $\kappa$ inferred within
          successive time windows of $\Delta{t}=0.1$, the SSE error
          Eq.~\ref{MSEgoal}, the scaling factor, and the frequency
          indicators associated with solving
          Eq.~\ref{para_reconstruct}, for different noise levels
          $\sigma^2$. Here, the SSE was minimized to find the estimate
          $\hat{\theta} \equiv \hat{\kappa}$ and the solutions ${u}_N$
          at intermediate timesteps $t_{j}+c_s\Delta{t}$.  (a, b)
          Smaller $\sigma^2$ leads to smaller SSE Eq.~\ref{MSEdef} and
          a more accurate reconstruction of $\hat{\kappa}$. When the
          function has spread out significantly at long times, the
          reconstructed ${\hat{\kappa}}$ becomes less accurate,
          suggesting that unboundedness and small function values
          render the problem susceptible to numerical difficulties.
          (c, d) Noisy data results in a larger proportion of
          high-frequency waves and thus a large frequency indicator,
          impeding proper scaling.}
     \label{fig6_lambda}
\end{figure}
Figure~\ref{fig6_lambda}(a) shows that the larger the noise, the less
accurate the reconstructed $\kappa$. Moreover, as the function becomes
more spread out (when $\sigma^{2}=0$), the error in both the
reconstructed diffusivity and the SSE increases across time, as shown
in Fig.~\ref{fig6_lambda}(b).  This behavior suggests that a diffusive
solution that decays more slowly at infinity can give rise to
inaccuracies in the numerical computation of the intermediate timestep
solutions and in reconstructing model parameters. Finally, as
indicated in Fig.~\ref{fig6_lambda}(c,d), larger variances in the
noise will impede the scaling process since the frequency indicator
cannot be as easily controlled because larger variance in the noise
usually corresponds to high-frequency and oscillatory components of a
solution.
\end{example}
%
%
In Example~\ref{KAPPA}, both the parameter and the unknown solution
were inferred.  Apart from reconstructing the coefficients in a given
physical model, in certain applications, we may also wish to
reconstruct the underlying physical model by inferring, \eg, the heat
source $f(x,t)$. Source recovery from observational data commonly
arises and has been the subject of many previous studies
\cite{yan2009meshless,yang2011inverse,yang2010simplified}.
We now discuss how the s-PINN methods presented here can also be used
for this purpose.  For example, in Eq.~\ref{parabolic} or
Eq.~\ref{para_reconstruct}, we may wish to reconstruct an unknown
source $f(x, t)$ by also approximating it with a spectral
decomposition
\begin{equation}
    f(x, t) \approx {f}_N(x, t)=\sum_{i=0}^N h_i(t)\phi_{i, x_L}^{\beta}(x),
    \label{freconstruct}
\end{equation}
and minimizing an SSE that is augmented by a penalty on the
coefficients ${h}_i, i=0,\ldots,N$.  

\noindent We learn the expansion coefficients ${h}_i$ within $[t_j,
  t_{j+1}]$ by minimizing
\begin{eqnarray}
{\rm SSE}_{j} = & {\rm SSE}_{j}^{\rm L} + {\rm SSE}_{j}^{\rm R}
  +\lambda\sum_{s=1}^K \big\|{\textbf{h}_N}(t_{j}+c_s\Delta{t})\big\|_2^2,\quad \lambda \geq 0, \nonumber
\end{eqnarray}

\begin{eqnarray}
{\rm SSE}_{j}^{\rm L}  =  & \sum_{s=1}^K \Big\| u(x, t_{j}+c_s\Delta{t}) 
- u(x, t_{j}) \nonumber\\[-5pt]
\: & \hspace{5mm} - \sum_{r=1}^K 
a_{sr}\big[\partial_{xx}u(x, t_{j}+c_r\Delta{t}) + {f}_N(x,t_{j}+c_r\Delta{t}; \Theta_{j+1})\big]\Big\|_{2}^{2}, \label{potentialloss} \\[3pt]
{\rm SSE}_{j}^{\rm R} =  & \sum_{s=1}^K \Big\| u(x, t_{j}+c_s\Delta{t}) 
- u(x, t_{j+1}) \nonumber\\[-5pt]
\: & \hspace{5mm} - \sum_{r=1}^K (a_{sr}-b_r)
\big[\partial_{xx}u(x, t_{j}+c_r\Delta{t}) + {f}_N(x,t_{j}+c_r\Delta{t}; 
\Theta_{j+1})\big]\Big\|_{2}^{2}, \nonumber
\end{eqnarray}
where ${\textbf{h}_N}(t_{j}+c_s\Delta{t})\equiv
({h}_1(t_{j}+c_s\Delta{t}),\ldots, {h}_{N}(t_{j}+c_s\Delta{t}))$ and 
$u$ (or the spectral expansion coefficients $w_{i}$ of
$u$) is assumed known at all intermediate time points
$c_{s}\Delta t$ in $(t_{j}, t_{j+1})$.

The last term in Eq.~\ref{potentialloss} adds an $L^2$ penalty term
on the coefficients of $f$ which tends to reconstruct smoother and
smaller-magnitude sources as $\lambda$ is increased.  Other forms of
regularization such as $L^1$ can also be considered
\cite{wu2019toward}. In the presence of noise, an $L^1$
regularization further drives small expansion weights to zero,
yielding an inferred source ${f}_N$ described by fewer nonzero
weights.

Since the reconstructed heat source $f_N$ is expressed in terms of a
spectral expansion in Eq.~\ref{freconstruct}, and minimizing the
loss function Eq.~\ref{potentialloss} depends on the global
information of the observation ${u}$, ${f}$ at any location $x$ also
contains global information intrinsic to ${u}$.  In other words, for
such inverse problems, the s-PINN approach extracts global spatial
information and is thus able to reconstruct global quantities.  We
consider an explicit case in the next example.

%
\begin{example}\hspace{-5pt}{\bf: Source recovery}
\label{example:potential}\\
\rm Consider the canonical source reconstruction problem
\cite{cannon1968determination,johansson2007variational,hasanov2014unified}
of finding $f(x, t)$ in the heat equation model in
Eq.~\ref{parabolic} for which observational data are given by
Eq.~\ref{noisy_data} but evaluated at $t_{j}+c_{s}\Delta t$.  A
physical interpretation of the reconstruction problem is identifying
the heat source $f(x, t)$ using measurement data in conjunction with
Eq.~\ref{parabolic}.  As in Example 5, we numerically solve the weak
form Eq.~\ref{para_weak}.  To study how the $L^2$ penalty term in
Eq.~\ref{potentialloss} affect source recovery and whether
increasing the regularization $\lambda$ will make the inference of $f$
more robust against noise, we minimize Eq.~\ref{MSEgoal} for
different values of $\lambda$ and $\sigma^2$.
\begin{table}[htb]
\centering
\caption{\small The error $\textrm{SSE}_0$ from Eq.~\ref{MSEdef} and
  the error of the reconstructed source Eq.~\ref{ferror}, under
  different strengths of data noise and regularization coefficients
  $\lambda$. The SSE is listed in the upper-right of each cell and the
  error of the reconstructed source (Eq.~\ref{ferror}) is listed in
  the lower-left of each cell.}
\vspace{2mm}
\scriptsize
\begin{tabular}{|l|R{2.8cm}|R{2.8cm}|R{2.8cm}|R{2.8cm}|}
\hline
\diagbox{$\sigma$}{$\lambda$} & 0 & $10^{-3}$ & $10^{-2}$ & $10^{-1}$ \\ 
\hline
0 &  \diagbox{0.1370}{1.543e-08} & \diagbox{\,\,\,0.1370\,\,\,}{1.368e-05} &
\diagbox{\,\,\,0.1477\,\,\,}{\,\,0.00132\,\,} & \diagbox{\,\,\,\,0.3228\,\,\,}{\,\,\,\,0.0888\,\,\,}\\
\hline $10^{-3}$ & \diagbox{0.1821}{2.837e-06} &
\diagbox{\,\,0.1818\,\,}{2.736e-05} & \diagbox{\,\,\,0.1702\,\,\,}{1.387e-03}&
\diagbox{\,\,\,\,0.3222\,\,\,}{\,0.08964\,}\\
\hline $10^{-2}$ & \diagbox{\,\,1.0497\,\,}{\,0.001517\,} &
\diagbox{\,\,\,1.0383\,\,\,}{1.579e-03} &
\diagbox{\,\,\,0.8031\,\,\,}{6.078e-03} &\diagbox{\,\,\,\,0.3434\,\,\,}{\,\,\,\,0.1168\,\,\,} \\
\hline $10^{-1}$ & \diagbox{\,\,\,11.505\,\,\,}{\,\,\,\,0.2976\,\,\,} &
\diagbox{\,\,\,11.458\,\,\,}{\,\,\,\,0.3032\,\,\,} &
\diagbox{\,\,\,8.2961\,\,\,}{\,\,\,\,0.6905\,\,\,} &\diagbox{\,\,\,\,1.3018\,\,\,}{\,\,\,\,2.9330\,\,\,}\\
\hline
    \end{tabular}%
\label{example7_table}
\end{table}

We use a neural network with 13 layers and 100 neurons per layer to
reconstruct $f_{i}(t)$ in the decomposition Eq.~\ref{freconstruct}
with $N=16$, \ie, the neural network outputs the coefficients $t{h}_i$
at the intermediate timesteps $t_j+c_s\Delta{t}$.  The basis functions
$\phi_{i, x_L}^{\beta}(x)$ are chosen to be Hermite functions
$\hat{\mathcal{H}}_{i, x_L}^{\beta}(x)$. For simplicity, we consider
the problem only at times within the first time point $[0, 0.2]$ and a
fixed scaling factor $\beta=0.8$ as well as a fixed displacement
$x_L=0$.

In Table \ref{example7_table}, we record the $L^2$ error 

\begin{equation}
    \big\|f(x, t) - \sum_{i=0}^{16} {h}_i(t)\hat{\mathcal{H}}_{i, x_L}^{\beta}(x)\big\|_{2}
\label{ferror}
\end{equation}
the lower-left of each entry and the ${\rm SSE}_{0}$ in the
upper-right. Observe that as the variance of the noise increases, the
reconstruction of $f$ via the spectral expansion becomes increasingly
inaccurate. 
 \begin{figure}[h!]
      \begin{center}
      \includegraphics[width=3in]{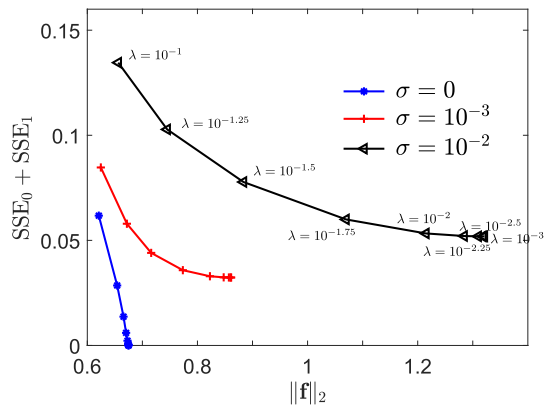}
	\end{center}
      \vspace{-4mm}
        \caption{\small Example~\ref{example:potential}: Source
            recovery. $\textrm{SSE}_0$ plotted against the
          reconstructed heat source $\|\textbf{h}_N\|_2$ as given by
          \ref{potentialloss}, as a function of $\lambda$ for
          various values of $\sigma^{2}$ (an ``L-curve'').  When
          $\lambda$ is large, the norm of the reconstructed heat
          source $\|{\textbf{h}_N}\|_2$ always tends to decrease while
          the ``error" $\textrm{SSE}_0$ tends to increase. When
          $\lambda=10^{-1}$, $\|{\textbf{h}_N}\|_2$ is small and the
          $\textrm{SSE}_0$ is large. A moderate $\lambda\in[10^{-2},
            10^{-3}]$ could reduce the error $\textrm{SSE}_0$,
          compared to using a large $\lambda$, while also generating a
          heat source with smaller $\|{\textbf{h}_N}\|_2$.}
     \label{fig7_lambda}
\end{figure}
In the noise-free case, taking $\lambda=0$ in Eq.~\ref{potentialloss}
achieves the smallest ${\rm SSE}_{0}$ and the smallest reconstruction
error. However, with increasing noise $\sigma^2$, using an $L^2$
regularization term in Eqs.~\ref{potentialloss} can prevent
over-fitting of the data although ${\rm SSE}_{0}$ increases with the
regularization strength $\lambda$.  When $\sigma=10^{-3}$, taking
$\lambda=10^{-2}$ achieves the smallest reconstruction error
Eq.~\ref{ferror}; when $\sigma=10^{-2}, 10^{-1}$, $\lambda=10^{-1}$
achieves the smallest reconstruction error. However, if $\lambda$ is
too large, coefficients of the spectral approximation to $f$ are
pushed to zero.  Thus, it is important to choose an intermediate
$\lambda$ so that the reconstruction of the source is robust to
noise. In Fig.~\ref{fig7_lambda}, we plot the norm of the
reconstructed heat source $\|{\textbf{h}_N}\|_2$ and the ``error"
$\textrm{SSE}_0$ which varies as $\lambda$ changes for different
$\sigma$.
\end{example}

%

\section{Summary and Conclusions}
\label{summary}
In this paper, we propose an approach that blends standard PINN
algorithms with adaptive spectral methods and show through examples
that this hybrid approach can be applied to a wide variety of
data-driven problems including function approximation, solving PDEs,
parameter inference, and model selection.  The underlying feature that
we exploit is the physical differences across classes of data. For
example, by understanding the difference between space and time
variables in a PDE model, we can describe the spatial dependence in
terms of basis functions, obviating the need to normalize spatial
data. Thus, s-PINNs are ideal for solving problems in unbounded
domains. The only additional ``prior'' needed is an assumption on the
asymptotic spatial behavior and an appropriate choice of basis
functions.  Additionally, adaptive techniques have been recently
developed to further improve the efficiency and accuracy, making
spectral decomposition especially suitable for unbounded-domain
problems that the standard PINN cannot easily address.

We applied s-PINNs (exploiting adaptive spectral methods) across a
number of examples and showed that they can outperform simple neural
networks for function approximation and existing PINNs for solving
certain PDEs. Three major advantages are that s-PINNs can be applied
to unbounded domain problems, more accurate by recovering spectral
convergence in space, and more efficient as a result of faster
evaluation of spatial derivatives of all orders compared to standard
PINNs that use autodifferentiation. These advantages are rooted in
separated data structures, allowing for spectral computation and
high-accuracy numerics. Straightforward implementation of s-PINNs
retains most of the advantageous features of deep-neural-network in
PINNs, making s-PINNs ideal for data-driven inference
problems. However, in the context of solving higher-dimensional PDEs,
a tradeoff is necessary when using s-PINNs instead of PINNs.  For
s-PINNs, the network structure needs to be significantly widened to
output an exponentially increasing (with dimensionality) number of
expansion coefficients, while in standard PINNs, the network structure
remains largely preserved but an exponentially larger number of
trajectories are needed for sufficient training. We found that by
restricting the spatial domain to a hyperbolic cross space, the number
of outputs required for s-PINNs can be appreciably decreased for
problems of moderate dimensions. While using a hyperbolic cross space
cannot reduce the number of outputs sufficiently to allow s-PINNs to
be effective for very high dimensional problems, the standard PINNs
approach to problems in very high dimensions could require an
unattainable number of samples for sufficient training.

In Table~\ref{tab:pros_cons}, we compare the advantages and
disadvantages of the standard PINN and s-PINN methods. Potential
improvements and extensions include applying techniques for selecting
basis functions that best characterize the expected underlying
process, spatial or otherwise, and inferring forms of the underlying
model PDEs \cite{long2018pde,raissi2018deep}. While standard PINN
methods deal with local information (\textit{e.g.}, $\partial_{x}u,
\partial_{x}^{2}u$), spectral decompositions capture global
information making them a natural choice for also efficiently learning
and approximating nonlocal terms such as convolutions and integral
kernels. Potential future exploration using our s-PINN method may
include adapting it to solve higher-dimensional problems by more
systematically choosing a proper hyperbolic space or using other
coefficient-reducing (outputs of the neural network) techniques. Also,
recent Gaussian\textendash process\textendash based smoothing
techniques \cite{bajaj2023recipes} can be considered to improve
robustness of our s-PINN method against noise/errors in measurements,
and noise-aware physics-informed machine learning techniques
\cite{thanasutives2022noise} can be incorporated when applying our
s-PINN for inverse-type PDE discovery problems.  Finally, one can
incorporate a recently proposed Bayesian-PINN (B-PINN)
\cite{yang2021b} method into our s-PINN method to quantify uncertainty
when solving inverse problems under noisy data.
\begin{landscape}
\begin{table}[p]
\footnotesize
\centering
\caption{Advantages and disadvantages of traditional and
    PINN-based numerical solvers. We use ``\pro'' and ``\con'' signs
  to indicate advantages and disadvantages, respectively. Finite
  difference (FD), finite-element (FE), and spectral methods can be
  used in a traditional sense without relying on neural networks. This
  table provides an overview of the advantages and disadvantages
  associated with the corresponding methods and solvers.}
\vspace{4mm}
\renewcommand*{\arraystretch}{1.2}
\begin{tabular}{|c|*{2}{l|}}\hline
\diagbox{\,\,\,{Methods}\,\,\,}{\,\,\,{Solvers}\,\,\,}
& \multicolumn{1}{c|}{\makebox[3em]{{Traditional}}} & \multicolumn{1}{c|}{\makebox[3em]{{PINN}}}  \\\hline\hline
{Non-spectral} & \,\,\,   
\makecell[l]{
\\
\pro~leverages existing numerical methods\\
\pro~low-order FD/FE schemes easily implemented\\
\pro~efficient evaluation of function and derivatives\\
\con~mainly restricted to bounded domains\\
\con~complicated time-extrapolation\\
\con~complicated implementation of higher-order schemes\\
\con~algebraic convergence, less accurate\\
\con~more complicated inverse-type problems\\
\con~more complicated temporal \\
\quad and spatial extrapolation\\
\con~requires understanding of problem to \\
\quad choose suitable discretization\\ \\}
 \,\, & \,\,\, 
\makecell[l]{
\\
\pro~easy implementation\\
\pro~efficient deep-neural-network training\\
\pro~easy extrapolation\\
\pro~easily handles inverse-type problems\\
\con~mainly restricted to bounded domains\\
\con~less accurate\\
\con~less interpretable spatial derivatives\\
\con~limited control of spatial discretization\\
\con~expensive evaluation of neural networks\\
\con~incompatible with existing numerical methods \\ \\} \,\, \\[8pt]\hline
Spectral & \,\,\,  \makecell[l]{
\pro~suitable for bounded and unbounded domains\\
\pro~spectral convergence in space, more accurate\\
\pro~leverage existing numerical methods\\
\pro~efficient evaluation of function and derivatives\\
\con~information required for choosing basis functions\\
\con~more complicated inverse-type problems\\
\con~more complicated implementation\\
\con~more complicated temporal extrapolation in time\\
\con~usually requires a ``regular" domain \\
\quad \textit{e.g.} rectangle, $\mathbb{R}^d$, a ball, etc.\\ \\} 
\,\, & \,\,\, 
\makecell[l]{
\\
\pro~suitable for both bounded and unbounded domains\\
\pro~easy implementation\\
\pro~spectral convergence in space, more accurate\\
\pro~efficient deep-neural-network training\\
\pro~more interpretable derivatives of spatial variables\\
\pro~easy extrapolation \\
\pro~easily handles inverse-type problems\\
\pro~compatible with existing adaptive techniques\\
\con~requires some information to choose basis functions\\
\con~expensive evaluation of neural networks\\
\con~usually requires a ``regular" domain \\
} \,\, \\[8pt]\hline
\end{tabular}
\label{tab:pros_cons}
\end{table}
\end{landscape}

\ack{LB acknowledges financial support from the Swiss National Fund
  (grant number P2EZP2\_191888). The authors also acknowledge support
  from the US Army Research Office (W911NF-18-1-0345) and the National
  Science Foundation (DMS-1814364).}

\bibliography{refs_new}

\begin{thebibliography}{10}

\bibitem{hornik1991approximation}
Kurt Hornik.
\newblock Approximation capabilities of multilayer feedforward networks.
\newblock {\em Neural Networks}, 4(2):251--257, 1991.

\bibitem{park2020minimum}
Sejun Park, Chulhee Yun, Jaeho Lee, and Jinwoo Shin.
\newblock Minimum width for universal approximation.
\newblock In {\em International Conference on Learning Representations}, 2020.

\bibitem{raissi2019physics}
Maziar Raissi, Paris Perdikaris, and George~E Karniadakis.
\newblock {P}hysics-informed neural networks: A deep learning framework for
  solving forward and inverse problems involving nonlinear partial differential
  equations.
\newblock {\em Journal of Computational Physics}, 378:686--707, 2019.

\bibitem{karniadakis2021physics}
George~Em Karniadakis, Ioannis~G Kevrekidis, Lu~Lu, Paris Perdikaris, Sifan
  Wang, and Liu Yang.
\newblock Physics-informed machine learning.
\newblock {\em Nature Reviews Physics}, 3(6):422--440, 2021.

\bibitem{asikis2020neural}
Thomas Asikis, Lucas B{\"o}ttcher, and Nino Antulov-Fantulin.
\newblock Neural ordinary differential equation control of dynamics on graphs.
\newblock {\em Physical Review Research (in press)}, 2022.

\bibitem{bottcher2021implicit}
Lucas B{\"o}ttcher, Nino Antulov-Fantulin, and Thomas Asikis.
\newblock {AI} {P}ontryagin or how neural networks learn to control dynamical
  systems.
\newblock {\em Nature Communications}, 2021.

\bibitem{bottcher2022near}
Lucas B{\"o}ttcher and Thomas Asikis.
\newblock Near-optimal control of dynamical systems with neural ordinary
  differential equations.
\newblock {\em Machine Learning: Science and Technology}, 3(4):045004, 2022.

\bibitem{lewis2020neural}
FW~Lewis, Suresh Jagannathan, and Aydin Yesildirak.
\newblock {\em Neural network control of robot manipulators and non-linear
  systems}.
\newblock CRC Press, 2020.

\bibitem{DBLP:journals/corr/abs-1710-10686}
Jan Kuka{\v{c}}ka, Vladimir Golkov, and Daniel Cremers.
\newblock Regularization for deep learning: A taxonomy.
\newblock {\em arXiv preprint arXiv:1710.10686}, 2017.

\bibitem{lutter2019deep}
M~Lutter, C~Ritter, and Jan Peters.
\newblock Deep {L}agrangian networks: Using physics as model prior for deep
  learning.
\newblock In {\em International Conference on Learning Representations}.
  OpenReview.net, 2019.

\bibitem{roehrl2020modeling}
Manuel~A Roehrl, Thomas~A Runkler, Veronika Brandtstetter, Michel Tokic, and
  Stefan Obermayer.
\newblock Modeling system dynamics with physics-informed neural networks based
  on {L}agrangian mechanics.
\newblock {\em IFAC-PapersOnLine}, 53(2):9195--9200, 2020.

\bibitem{zhong2019symplectic}
Yaofeng~Desmond Zhong, Biswadip Dey, and Amit Chakraborty.
\newblock Symplectic {ODE}-net: Learning {H}amiltonian dynamics with control.
\newblock In {\em International Conference on Learning Representations}, 2019.

\bibitem{kharazmi2019variational}
Ehsan Kharazmi, Zhongqiang Zhang, and George~Em Karniadakis.
\newblock Variational physics-informed neural networks for solving partial
  differential equations.
\newblock {\em arXiv preprint arXiv:1912.00873}, 2019.

\bibitem{jagtap2020extended}
Ameya~D Jagtap and George~Em Karniadakis.
\newblock Extended physics-informed neural networks (xpinns): A generalized
  space-time domain decomposition based deep learning framework for nonlinear
  partial differential equations.
\newblock {\em Communications in Computational Physics}, 28(5):2002--2041,
  2020.

\bibitem{li2021physics}
Zongyi Li, Hongkai Zheng, Nikola Kovachki, David Jin, Haoxuan Chen, Burigede
  Liu, Kamyar Azizzadenesheli, and Anima Anandkumar.
\newblock Physics-informed neural operator for learning partial differential
  equations.
\newblock {\em arXiv preprint arXiv:2111.03794}, 2021.

\bibitem{mao2020physics}
Zhiping Mao, Ameya~D Jagtap, and George~Em Karniadakis.
\newblock Physics-informed neural networks for high-speed flows.
\newblock {\em Computer {M}ethods in {A}pplied {M}echanics and {E}ngineering},
  360:112789, 2020.

\bibitem{fang2019physics}
Zhiwei Fang and Justin Zhan.
\newblock A physics-informed neural network framework for {PDE}s on 3{D}
  surfaces: Time independent problems.
\newblock {\em IEEE Access}, 8:26328--26335, 2019.

\bibitem{misyris2020physics}
George~S Misyris, Andreas Venzke, and Spyros Chatzivasileiadis.
\newblock Physics-informed neural networks for power systems.
\newblock In {\em 2020 IEEE Power \& Energy Society General Meeting (PESGM)},
  pages 1--5. IEEE, 2020.

\bibitem{sahli2020physics}
Francisco Sahli~Costabal, Yibo Yang, Paris Perdikaris, Daniel~E Hurtado, and
  Ellen Kuhl.
\newblock Physics-informed neural networks for cardiac activation mapping.
\newblock {\em Frontiers in Physics}, 8:42, 2020.

\bibitem{liu2020generic}
Minliang Liu, Liang Liang, and Wei Sun.
\newblock A generic physics-informed neural network-based constitutive model
  for soft biological tissues.
\newblock {\em Computer methods in applied mechanics and engineering},
  372:113402, 2020.

\bibitem{bottcher2021computational}
Lucas B{\"o}ttcher and Hans~J Herrmann.
\newblock {\em Computational Statistical Physics}.
\newblock Cambridge University Press, 2021.

\bibitem{strub2019modeling}
Stefan~H Strub and Lucas B{\"o}ttcher.
\newblock Modeling deformed transmission lines for continuous strain sensing
  applications.
\newblock {\em Measurement Science and Technology}, 31(3):035109, 2019.

\bibitem{barre2011algebraic}
Julien Barr{\'e}, Alain Olivetti, and Yoshiyuki~Y Yamaguchi.
\newblock {Algebraic damping in the one-dimensional {V}lasov equation}.
\newblock {\em Journal of Physics A: Mathematical and Theoretical},
  44(40):405502, 2011.

\bibitem{li2018stability}
Buyang Li, Jiwei Zhang, and Chunxiong Zheng.
\newblock Stability and error analysis for a second-order fast approximation of
  the one-dimensional {Schr\"odinger} equation under absorbing boundary
  conditions.
\newblock {\em SIAM Journal on Scientific Computing}, 40(6):A4083--A4104, 2018.

\bibitem{xia2020pde}
Mingtao Xia, Chris~D Greenman, and Tom Chou.
\newblock {PDE} models of adder mechanisms in cellular proliferation.
\newblock {\em SIAM Journal on Applied Mathematics}, 80(3):1307--1335, 2020.

\bibitem{xia2021kinetic}
Mingtao Xia and Tom Chou.
\newblock Kinetic theory for structured populations: application to stochastic
  sizer-timer models of cell proliferation.
\newblock {\em Journal of Physics A: Mathematical and Theoretical}, 2021.

\bibitem{mengotti2011real}
Elena Mengotti, Laura~J Heyderman, Arantxa~Fraile Rodr{\'\i}guez, Frithjof
  Nolting, Remo~V H{\"u}gli, and Hans-Benjamin Braun.
\newblock {Real-space observation of emergent magnetic monopoles and associated
  {D}irac strings in artificial {K}agom\'e spin ice}.
\newblock {\em Nature Physics}, 7(1):68--74, 2011.

\bibitem{hugli2012artificial}
RV~H{\"u}gli, G~Duff, B~O'Conchuir, E~Mengotti, A~Fraile Rodr{\'\i}guez,
  F~Nolting, LJ~Heyderman, and HB~Braun.
\newblock {Artificial {K}agom\'e spin ice: dimensional reduction, avalanche
  control and emergent magnetic monopoles}.
\newblock {\em Philosophical Transactions of the Royal Society A: Mathematical,
  Physical and Engineering Sciences}, 370(1981):5767--5782, 2012.

\bibitem{xia2021efficient}
Mingtao Xia, Sihong Shao, and Tom Chou.
\newblock Efficient scaling and moving techniques for spectral methods in
  unbounded domains.
\newblock {\em SIAM Journal on Scientific Computing}, 43(5):A3244--A3268, 2021.

\bibitem{xia2021frequency}
Mingtao Xia, Sihong Shao, and Tom Chou.
\newblock A frequency-dependent p-adaptive technique for spectral methods.
\newblock {\em Journal of Computational Physics}, 446:110627, 2021.

\bibitem{shen2011spectral}
Jie Shen, Tao Tang, and Li-Lian Wang.
\newblock {\em Spectral methods: algorithms, analysis and applications},
  volume~41.
\newblock Springer Science \& Business Media, 2011.

\bibitem{trefethen2000spectral}
Lloyd~N Trefethen.
\newblock {\em Spectral methods in MATLAB}.
\newblock SIAM, 2000.

\bibitem{linnainmaa1976taylor}
Seppo Linnainmaa.
\newblock Taylor expansion of the accumulated rounding error.
\newblock {\em BIT Numerical Mathematics}, 16(2):146--160, 1976.

\bibitem{paszke2017automatic}
Adam Paszke, Sam Gross, Soumith Chintala, Gregory Chanan, Edward Yang, Zachary
  DeVito, Zeming Lin, Alban Desmaison, Luca Antiga, and Adam Lerer.
\newblock Automatic differentiation in {P}y{T}orch.
\newblock 2017.

\bibitem{hornik1989multilayer}
Kurt Hornik, Maxwell Stinchcombe, and Halbert White.
\newblock Multilayer feedforward networks are universal approximators.
\newblock {\em Neural Networks}, 2(5):359--366, 1989.

\bibitem{ioffe2015batch}
Sergey Ioffe and Christian Szegedy.
\newblock Batch normalization: Accelerating deep network training by reducing
  internal covariate shift.
\newblock In {\em {I}nternational Conference on Machine Learning}, pages
  448--456. PMLR, 2015.

\bibitem{tang2020rational}
Tao Tang, Li-Lian Wang, Huifang Yuan, and Tao Zhou.
\newblock {Rational spectral methods for PDEs involving fractional Laplacian in
  unbounded domains}.
\newblock {\em SIAM Journal on Scientific Computing}, 42(2):A585--A611, 2020.

\bibitem{baydin2018automatic}
Atilim~Gunes Baydin, Barak~A Pearlmutter, Alexey~Andreyevich Radul, and
  Jeffrey~Mark Siskind.
\newblock Automatic differentiation in machine learning: a survey.
\newblock {\em Journal of Machine Learning Research}, 18, 2018.

\bibitem{brandstetter2021message}
Johannes Brandstetter, Daniel~E Worrall, and Max Welling.
\newblock Message passing neural {PDE} solvers.
\newblock In {\em International Conference on Learning Representations}, 2021.

\bibitem{li2020fourier}
Zongyi Li, Nikola~Borislavov Kovachki, Kamyar Azizzadenesheli, Kaushik
  Bhattacharya, Andrew Stuart, Anima Anandkumar, et~al.
\newblock Fourier neural operator for parametric partial differential
  equations.
\newblock In {\em International Conference on Learning Representations}, 2020.

\bibitem{shen2010sparse}
Jie Shen and Li-Lian Wang.
\newblock Sparse spectral approximations of high-dimensional problems based on
  hyperbolic cross.
\newblock {\em SIAM Journal on Numerical Analysis}, 48(3):1087--1109, 2010.

\bibitem{pmlr-v80-arora18a}
Sanjeev Arora, Nadav Cohen, and Elad Hazan.
\newblock On the optimization of deep networks: {I}mplicit acceleration by
  overparameterization.
\newblock In Jennifer Dy and Andreas Krause, editors, {\em Proceedings of the
  35th International Conference on Machine Learning}, volume~80 of {\em
  Proceedings of Machine Learning Research}, pages 244--253, 2018.

\bibitem{chen2020much}
Zixiang Chen, Yuan Cao, Difan Zou, and Quanquan Gu.
\newblock How much over-parameterization is sufficient to learn deep {ReLU}
  networks?
\newblock In {\em International Conference on Learning Representations}, 2020.

\bibitem{huntul2021identification}
Mousa~J. Huntul.
\newblock Identification of the timewise thermal conductivity in a 2{D} heat
  equation from local heat flux conditions.
\newblock {\em Inverse Problems in Science and Engineering}, 29(7):903--919,
  2021.

\bibitem{ivanchov1993inverse}
NI~Ivanchov.
\newblock Inverse problems for the heat-conduction equation with nonlocal
  boundary conditions.
\newblock {\em Ukrainian Mathematical Journal}, 45(8):1186--1192, 1993.

\bibitem{jones1962determination}
B~Frank Jones~Jr.
\newblock The determination of a coefficient in a parabolic differential
  equation: Part i. existence and uniqueness.
\newblock {\em Journal of Mathematics and Mechanics}, pages 907--918, 1962.

\bibitem{beznoshchenko1974on}
N.~Ya. Beznoshchenko.
\newblock On finding a coefficient in a parabolic equation.
\newblock {\em Differential Equations}, 10:24--35, 1974.

\bibitem{yan2009meshless}
Liang Yan, Feng-Lian Yang, and Chu-Li Fu.
\newblock A meshless method for solving an inverse spacewise-dependent heat
  source problem.
\newblock {\em Journal of Computational Physics}, 228(1):123--136, 2009.

\bibitem{yang2011inverse}
Liu Yang, Mehdi Dehghan, Jian-Ning Yu, and Guan-Wei Luo.
\newblock Inverse problem of time-dependent heat sources numerical
  reconstruction.
\newblock {\em Mathematics and Computers in Simulation}, 81(8):1656--1672,
  2011.

\bibitem{yang2010simplified}
Fan Yang and Chu-Li Fu.
\newblock {A simplified {T}ikhonov regularization method for determining the
  heat source}.
\newblock {\em Applied Mathematical Modelling}, 34(11):3286--3299, 2010.

\bibitem{wu2019toward}
Tailin Wu and Max Tegmark.
\newblock Toward an artificial intelligence physicist for unsupervised
  learning.
\newblock {\em Physical Review E}, 100(3):033311, 2019.

\bibitem{cannon1968determination}
John~Rozier Cannon.
\newblock Determination of an unknown heat source from overspecified boundary
  data.
\newblock {\em SIAM Journal on Numerical Analysis}, 5(2):275--286, 1968.

\bibitem{johansson2007variational}
B~Tomas Johansson and Daniel Lesnic.
\newblock A variational method for identifying a spacewise-dependent heat
  source.
\newblock {\em IMA Journal of Applied Mathematics}, 72(6):748--760, 2007.

\bibitem{hasanov2014unified}
Alemdar Hasanov and Burhan Pekta{\c{c}}.
\newblock A unified approach to identifying an unknown spacewise dependent
  source in a variable coefficient parabolic equation from final and integral
  overdeterminations.
\newblock {\em Applied Numerical Mathematics}, 78:49--67, 2014.

\bibitem{long2018pde}
Zichao Long, Yiping Lu, Xianzhong Ma, and Bin Dong.
\newblock {PDE}-net: Learning {PDE}s from data.
\newblock In {\em International Conference on Machine Learning}, pages
  3208--3216. PMLR, 2018.

\bibitem{raissi2018deep}
Maziar Raissi.
\newblock Deep hidden physics models: {D}eep learning of nonlinear partial
  differential equations.
\newblock {\em Journal of Machine Learning Research}, 19(1):932--955, 2018.

\bibitem{bajaj2023recipes}
Chandrajit Bajaj, Luke McLennan, Timothy Andeen, and Avik Roy.
\newblock Recipes for when physics fails: Recovering robust learning of physics
  informed neural networks.
\newblock {\em Machine Learning: Science and Technology}, 2023.

\bibitem{thanasutives2022noise}
Pongpisit Thanasutives, Takashi Morita, Masayuki Numao, and Ken-ichi Fukui.
\newblock Noise-aware physics-informed machine learning for robust pde
  discovery.
\newblock {\em Machine Learning: Science and Technology}, 4:015009, 2022.

\bibitem{yang2021b}
Liu Yang, Xuhui Meng, and George~Em Karniadakis.
\newblock {B-PINNs: Bayesian physics-informed neural networks for forward and
  inverse PDE problems with noisy data}.
\newblock {\em Journal of Computational Physics}, 425:109913, 2021.

\end{thebibliography}
\bibliographystyle{unsrt.bst}

\end{document}